\documentclass[11pt,a4paper,copyright]{google}

\usepackage[authoryear,sort&compress,round]{natbib}
\usepackage{mathtools}
\usepackage{array}
\usepackage{longtable}
\usepackage{algorithm}
\usepackage{algorithmic}
\usepackage{placeins}
\usepackage{wrapfig}
\usepackage{xurl}

\newcolumntype{L}[1]{>{\raggedright\arraybackslash}p{#1}}

\renewcommand{\floatpagefraction}{0.7}
\newcommand{\cmark}{\textcolor{black}{\ding{51}}}
\newcommand{\xmark}{\textcolor{red}{\ding{55}}}
\definecolor{algdirbg}{RGB}{252,231,229}
\definecolor{algcreditbg}{RGB}{229,239,251}
\definecolor{algallocbg}{RGB}{255,240,214}
\definecolor{backbonerow}{gray}{0.94}
\makeatletter
\newcommand{\shadebackbonerow}{\noalign{%
  \begingroup\color{backbonerow}%
  \hrule height\dimexpr\ht\@arstrutbox+\dp\@arstrutbox\relax
  \endgroup
  \vskip-\dimexpr\ht\@arstrutbox+\dp\@arstrutbox\relax}}
\makeatother
\newcommand{\algcomment}[1]{\hfill\mbox{\footnotesize\color{black}\(\triangleright\)~#1}}
\newcommand{\alghl}[2]{\begingroup\setlength{\fboxsep}{1.2pt}%
  \colorbox{#1}{\strut\ensuremath{\color{black}#2}}\endgroup}

\newenvironment{alignedEquation}[1]{%
  \begin{equation}\label{#1}%
  \begin{aligned}%
}{%
  \end{aligned}%
  \end{equation}%
}

\renewcommand{\absfont}{\normalfont\linespread{1.2}\fontsize{11}{12}\selectfont}
\definecolor{royalblue}{RGB}{65,105,225}
\usepackage[
  colorlinks=true,
  linkcolor=red,
  urlcolor=royalblue,
  citecolor=brown,
  anchorcolor=blue
]{hyperref} \usepackage{ravec_frontmatter}

\newcommand{\papertitle}{RAVEC: Reward-Aligned Visual Evidence Correction}
\title{\papertitle}
\renewcommand{\Affilfont}{\normalfont\fontsize{9.2}{12}\selectfont}

\author[1]{Junjie~Yang}
\author[2\,\CorrespondingAuthorSymbol]{Rui~Qian}
\author[3]{Xutao~Mao}
\author[4]{Yuhe~Wu}
\author[5]{Mingda~Zhang}
\author[6]{Jiaxuan~Luo}
\author[7]{Yang~Tan}
\author[2]{Qiang~Huang}
\author[8]{Zhisheng~Chen}
\author[9]{Yanfeng~Li}
\author[10]{Dawen~Tian}
\author[11]{Changmiao~Wang}
\author[9]{Tao~Tan}
\affil[1]{\mbox{University of Maryland, College Park}}
\affil[2]{\mbox{Fudan University}}
\affil[3]{\mbox{City University of Hong Kong}}
\affil[4]{\mbox{The Hong Kong University of Science and Technology Guangzhou}}
\affil[5]{\mbox{The Chinese University of Hong Kong Shenzhen}}
\affil[6]{\mbox{Johns Hopkins University}}
\affil[7]{\mbox{Shanghai Innovation Institute}}
\affil[8]{\mbox{University of the Chinese Academy of Sciences}}
\affil[9]{\mbox{Macao Polytechnic University}}
\affil[10]{\mbox{Xiangtan University}}
\affil[11]{\mbox{Shenzhen Research Institute of Big Data}}

\begin{abstract}
Reinforcement learning with verifiable rewards improves multimodal
reasoning, but response-level feedback provides limited guidance on
which intermediate decisions to correct.
Visual interventions reveal sensitivity to image content, yet this
sensitivity alone does not establish reward alignment or support
across questions.
We introduce \textbf{RAVEC} (Reward-Aligned Visual Evidence Correction),
which selects visual corrections using local and cross-question reward
support and jointly determines their strengths within each response.
Paired Visual Evidence Contrast (PVEC) compares predictions from an
exponential moving average teacher at the same sampled response prefix
under the original image and a version with its annotated evidence
region suppressed, yielding signed correction directions.
Credit--Risk Allocation (CRA) screens these directions using local
reward alignment and reward support from other questions, excluding
all responses to the source question.
It then solves a response-level budgeted allocation problem, assigning
additional cost to directions that locally reduce predictive entropy.
The resulting signed visual loss augments Group Relative Policy
Optimization (GRPO) during training; inference uses only the original
image and question.
Across ten multimodal benchmarks, RAVEC achieves the highest average
accuracy among the compared methods with both 3B and 7B
Qwen2.5-VL-Instruct models, outperforming GRPO by 2.67 and 1.39
percentage points, respectively.
Component ablations support the combined design of evidence alignment,
credit qualification, and risk-aware allocation.
\end{abstract}

\begin{document}
\maketitle
\raggedbottom
\interlinepenalty=10000

\section{Introduction}
\label{sec:intro}

Multimodal reasoning requires models to connect visual observations to
the intermediate decisions that lead to an answer~\citep{mathverse2024}. A misread quantity in
a table or an overlooked relation in a diagram can invalidate an otherwise
coherent reasoning chain~\citep{mathvista2024,mathverse2024}.
Reinforcement learning with verifiable rewards~\citep{ref15,ref16,ref56,ref83}
improves this capability through feedback on complete answers.
Group Relative Policy Optimization (GRPO)~\citep{ref15}, however, assigns
the same response-level advantage to every generated token.
This shared signal provides limited guidance for locating visual errors
or distinguishing them from errors in the subsequent reasoning steps.
Learning from visual evidence therefore requires a more selective way
to determine which decisions to correct and how strongly to update them.

\begin{wraptable}[16]{R}{0.41\linewidth}\vspace{-12pt}
\centering
\captionsetup{width=\linewidth,font=footnotesize,justification=raggedright,singlelinecheck=false}
\caption{Evidence guidance and credit allocation.
EC denotes an evidence contrast signal.
TL denotes explicit token-level selection or weighting.
CS denotes scaling by directional cost, excluding ordinary KL regularization.
PC denotes credit support from other prompts.
JA denotes joint token weight allocation under a response budget.}
\label{tab:mechanism}
\footnotesize
\newcommand{\methodcite}[1]{{\scriptsize\citep{#1}}}
\setlength{\tabcolsep}{2pt}
\renewcommand{\arraystretch}{1.08}
\setlength{\aboverulesep}{1.5pt}
\setlength{\belowrulesep}{1.5pt}
\begin{tabular*}{\linewidth}{@{\extracolsep{\fill}}lccccc@{}}
\toprule
Method & EC & TL & CS & PC & JA \\
\midrule
GRPO~\methodcite{ref15} & \xmark & \xmark & \xmark & \xmark & \xmark \\
PAPO~\methodcite{ref56} & \cmark & \xmark & \xmark & \xmark & \xmark \\
CFPO~\methodcite{ref80} & \cmark & \xmark & \xmark & \xmark & \xmark \\
CPPO~\methodcite{ref77} & \cmark & \cmark & \xmark & \xmark & \xmark \\
PRPO~\methodcite{ref78} & \cmark & \cmark & \xmark & \xmark & \xmark \\
TPAE~\methodcite{ref81} & \cmark & \cmark & \xmark & \xmark & \xmark \\
SIVA-RL~\methodcite{ref82} & \cmark & \xmark & \xmark & \xmark & \xmark \\
VERPO~\methodcite{verpo2026} & \cmark & \cmark & \cmark & \xmark & \xmark \\
\midrule
\textbf{RAVEC (ours)} & \cmark & \cmark & \cmark & \cmark & \cmark \\
\bottomrule
\end{tabular*}
\end{wraptable}
Process supervision~\citep{ref21,ref22,ref23} provides feedback on intermediate
steps. Teacher replay~\citep{ref35,ref58} offers another route to
token-level guidance by evaluating predictions at sampled response
prefixes.
Visual interventions complement teacher guidance by comparing
predictions under altered image content~\citep{vcd2024}.
Table~\ref{tab:mechanism} compares how existing methods use this information.

PAPO~\citep{ref56} and CFPO~\citep{ref80} introduce evidence contrasts (EC)
through visual perturbations; CPPO~\citep{ref77}, PRPO~\citep{ref78}, and
TPAE~\citep{ref81} additionally select or weight individual tokens (TL).
These mechanisms help locate decisions that are sensitive to visual
input, but sensitivity alone does not establish agreement with the reward
objective. A teacher may react to degraded image quality or change its
confidence even when the resulting shift is unhelpful for the task.
The contrast therefore proposes a correction direction whose magnitude
alone cannot determine whether or how strongly the actor should follow it.

Two limitations remain in the prior methods summarized in Table~\ref{tab:mechanism}:
correction weights lack cross-question reward support and joint allocation
within a response.
First, corrections act through shared model parameters, so agreement with
the reward direction of one response does not imply agreement with
updates from other questions. Responses to the same question also share
its image and reward context. A peer signal should therefore come from
other questions to assess support beyond the source question.
Second, independent token weights leave the aggregate correction cost
uncoordinated. Corrections of equal magnitude may differ in reward support
and effects on predictive uncertainty. Accepting them individually does
not determine how their strengths should be distributed within a response.
These considerations motivate combining cross-question reward support
with joint allocation that accounts for predictive uncertainty.

\begin{figure}[!t]
  \centering
  \includegraphics[width=\linewidth]{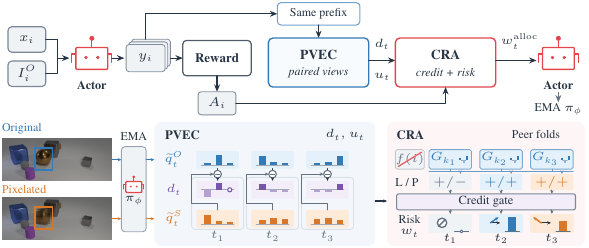}
  \input{figures/RAVEC_main_caption}
  \label{fig:ravec_overview}
\end{figure}

We introduce \textbf{RAVEC}, \textbf{\underline{R}}eward-\textbf{\underline{A}}ligned
\textbf{\underline{V}}isual \textbf{\underline{E}}vidence \textbf{\underline{C}}orrection,
with two complementary modules, as shown in Figure~\ref{fig:ravec_overview}.
Paired Visual Evidence Contrast (PVEC) replays an exponential moving
average (EMA) teacher at the same sampled response prefix using the
original image and a version with its annotated evidence region pixelated.
Reference answers guide region annotation during training, but neither
the actor nor the replay teacher receives those answers.
Inference uses only the original image and question.
With the teacher, question, and prefix fixed, the signed contrast proposes
which token preferences to strengthen or suppress.

Credit--Risk Allocation (CRA) qualifies each proposal using local reward
alignment and peer support. It maps the proposed logit correction to the
shared output head, where its induced update is compared with reward
gradients from other questions, excluding the entire fold containing the
source question and all its responses. Only proposals with positive local
and peer support receive weight. CRA then maximizes total credit-weighted correction strength under
a response energy budget, with directional cost accounting for magnitude
and local entropy contraction. The allocation has a closed-form solution.

Paired replay, response-level rewards, and region annotations thus supply
token-level guidance without correctness labels for intermediate reasoning
steps or a separately trained process verifier.
The signed visual loss augments GRPO, while an original-view EMA
reference provides separate policy regularization.

\Needspace{5\baselineskip}
We make three contributions:

\begin{itemize}
    \item We formulate visual evidence correction as a joint problem of
    cross-question credit qualification and response-level allocation.
    This separates local reward alignment from cross-question support
    and couples accepted corrections through a shared response budget.
    \item We present RAVEC, pairing signed PVEC proposals with credit
    qualification and joint allocation in CRA.
    CRA combines
    local alignment with cross-question output-head reward support, then
    allocates strengths in closed form under a response budget
    with a local entropy-contraction penalty.
    \item We evaluate RAVEC on ten multimodal benchmarks with 3B and 7B
    Qwen2.5-VL-Instruct~\citep{qwen25vl2025} models, achieving the highest average accuracy among the
    compared methods and outperforming GRPO on every benchmark at both
    scales. Component ablations support combining evidence alignment with credit qualification and risk-aware allocation. Controlled training analyses track learning dynamics; area-matched controls test sensitivity to region choice.
\end{itemize}

\begingroup
\makeatletter
\renewcommand{\paragraph}{\@startsection{paragraph}{4}{\z@}
  {4pt}{-1em}{\normalfont\normalsize\bfseries}}
\makeatother
\section{Related Work}
\label{sec:related}

\paragraph{Visual Reinforcement Learning and Interventions.}
Visual-RFT~\citep{visualrft2025} applies verifiable rewards to perception;
DeepEyes~\citep{deepeyes2026} learns active perception through reinforcement
learning. Vision-R1~\citep{ref83} uses multimodal reasoning traces for
cold starts and progressively suppresses overthinking during RL.
Visual interventions provide complementary signals about how
predictions depend on image content.
Visual Contrastive Decoding~\citep{vcd2024} contrasts original and distorted
image predictions during decoding.
PAPO~\citep{ref56} regularizes KL and entropy under image masking,
while CFPO~\citep{ref80} regularizes contrasts under salient visual interventions.
CPPO~\citep{ref77} supervises positively advantaged perceptual tokens.
SIVA-RL~\citep{ref82} routes visual pairs toward sensitivity
or invariance using reward changes.
RAVEC contrasts paired teacher predictions at fixed response prefixes
to propose signed corrections, then selects and jointly weights them using
reward support. These corrections affect training only; inference uses
the original input without contrasting predictions.

\paragraph{Local Credit and Selective Supervision.}
Gradient agreement guides validation-based example reweighting~\citep{reweight2018}
and TracIn~\citep{tracin2020} training influence estimation.
PCGrad~\citep{pcgrad2020} projects conflicting task gradients to reduce
interference during shared-parameter optimization.
Process supervision~\citep{ref21,ref22,ref23} verifies intermediate steps.
PRPO~\citep{ref78} adjusts token advantages using visual dependence and
perturbation stability. TPAE~\citep{ref81} uses correct rollouts as references
for visual dependence and entropy.
RAVEC qualifies each correction using local reward alignment and
cross-question output-head support, excluding all responses to the source
question. This proxy selects proposals without projecting full task gradients.

\paragraph{Self-Distillation and Correction Control.}
On-policy distillation~\citep{ref35} supervises student trajectories.
MiniLLM~\citep{ref36} minimizes reverse KL to transfer a larger
teacher's distribution.
OPSD~\citep{ref58} and SDPO~\citep{ref57} enrich replay with solutions and
feedback to guide self-distillation.
SRPO~\citep{ref59} routes samples between reward optimization and self-distillation.
RLCSD~\citep{ref61} and CEPO~\citep{ref62} select supervision from evidence contrasts.
VERPO~\citep{verpo2026} accepts and scales corrections using local reward
alignment and Fisher movement cost.
RAVEC combines credit qualification with joint allocation under a shared
response budget, using each correction's credit, magnitude, and local
entropy-contraction penalty. Its visual loss uses signed evidence contrasts, while
a separate EMA KL term provides distributional regularization.
\par
\endgroup

\clearpage

\section{Proposed RAVEC}
\label{sec:method}

RAVEC turns paired visual predictions into signed correction proposals,
qualifies them using local and cross-question reward support, and jointly
allocates their strengths within each response, as shown in Figure~\ref{fig:ravec_overview}.
PVEC constructs the directions; CRA selects and weights them.

\subsection{Problem Setup and Outcome Supervision}
\label{sec:grpo}

For sampled response $y_i$ of length $T_i$, let $x_i,I_i,\mathcal R_i$
denote its question, image, and annotated evidence region. The original
view is $I_i^O=I_i$; the suppressed view
$I_i^S=\mathcal T(I_i;\mathcal R_i)$ pixelates that region.
Region annotation uses the reference answer during training, but neither
the actor nor the replay teacher receives the answer or annotation reasoning.
The actor samples from the original view, with token distribution
$p_{\theta,i,t}=\pi_\theta(\cdot\mid x_i,I_i^O,y_{i,<t})$ over vocabulary
$\mathcal V$. Binary masks $m_{i,t}$ and $r_i$ indicate valid tokens and
available evidence regions, respectively.

GRPO~\citep{ref15} compares verifier rewards $R_i$ within each
same-question response group $\mathcal G(i)$, using
$A_i=(R_i-\bar R_{\mathcal G(i)})/(\sigma_{\mathcal G(i)}+10^{-6})$,
where $\bar R$ and $\sigma$ are the group mean and sample standard deviation.
We use the clipped objective~\citep{ref4}, $\mathcal L_{\mathrm{GRPO}}$,
averaged over $N_{\mathrm{tok}}=\sum_{i,t}m_{i,t}$ tokens~\citep{ref17}.
The shared $A_i$ supplies outcome supervision; later stages
add token-specific visual corrections.

\Needspace{4\baselineskip}
\subsection{Paired Visual Evidence Contrast}
\label{sec:pvec}

\textbf{Paired Replay.} At rollout start, an EMA teacher~\citep{mean_teacher2017}
produces $q^e_{i,t}=\pi_\phi(\cdot\mid x_i,I_i^e,y_{i,<t})$ for
$e\in\{O,S\}$.

At each position $t$, the two branches share the teacher,
question, and sampled prefix; only the annotated region is changed. This
pairing provides evidence alignment (EA).

Given the original view and sampled prefix, the actor produces logits $z_{i,t}$ at rollout start.
Equation~\eqref{eq:pvec_direction} uses the corresponding next-token distribution
$p_{i,t}=\operatorname{softmax}(z_{i,t})$ to turn the teacher contrast into a correction to those logits.

\Needspace{14\baselineskip}
\begin{wrapfigure}{r}{0.52\linewidth}
  \centering
  \captionsetup{font=small,justification=raggedright,skip=4pt}
  \includegraphics[width=\linewidth]{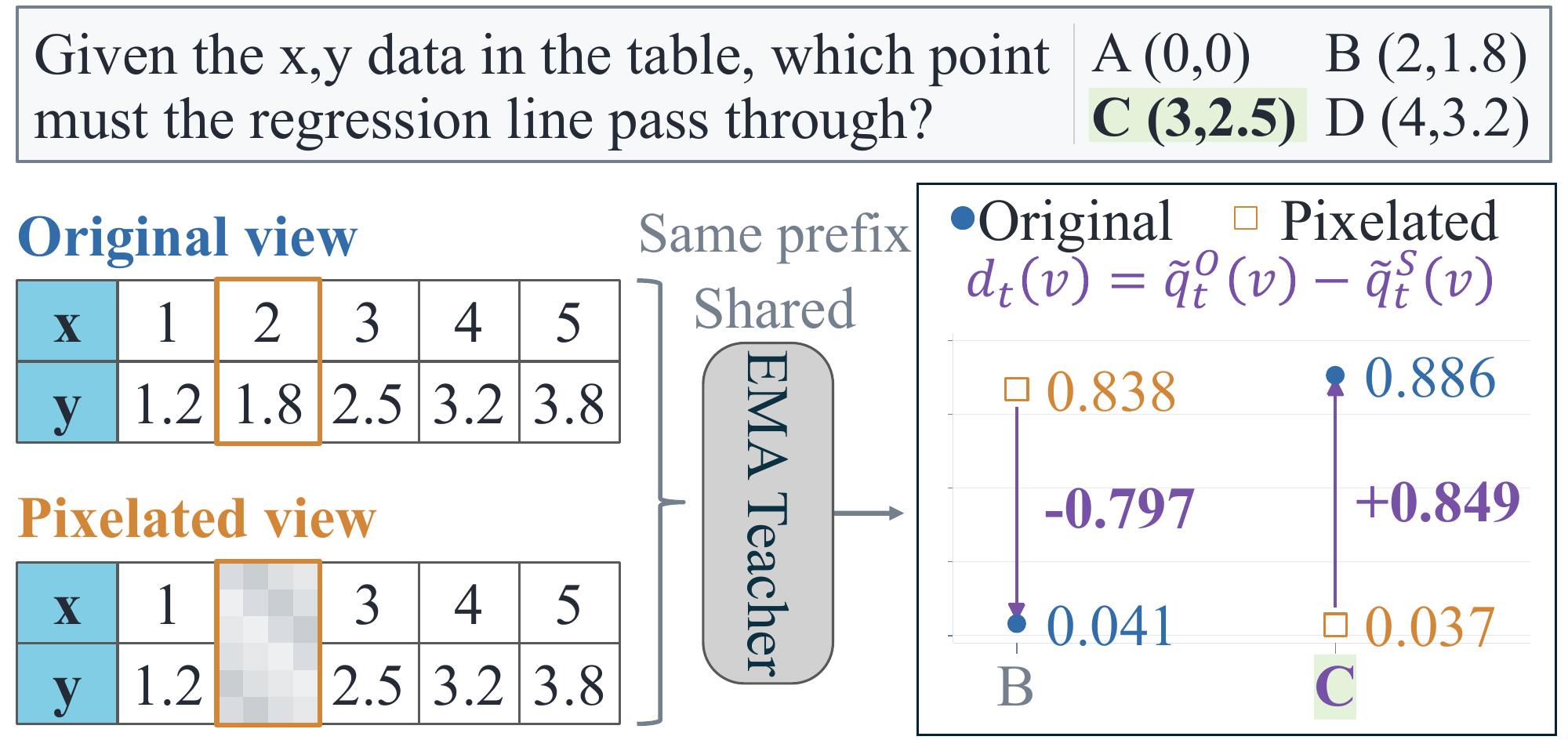}
  \caption{\textbf{PVEC: proposing a direction.}
    At the same prefix, the original view favors C and suppression favors B,
    giving opposite-signed coefficients in Eq.~\eqref{eq:pvec_direction}.}
  \label{fig:pvec_mechanism}
\end{wrapfigure}
Given the original view and sampled prefix, the actor produces logits $z_{i,t}$ at rollout start.
Equation~\eqref{eq:pvec_direction} uses the corresponding next-token distribution
$p_{i,t}=\operatorname{softmax}(z_{i,t})$ to turn the teacher contrast into a correction to those logits.

Let $\mathcal S=\operatorname{Top}_K(q^O)\cup \operatorname{Top}_K(q^S)\cup\{y\}$, where $y$ denotes $y_{i,t}$.
For each endpoint, soften probabilities~\citep{ref33} on $\mathcal S$ at
temperature $\tau_v=2$, preserving its retained mass
$M_e=\sum_{v\in\mathcal S}q^e(v)$ and setting other entries to zero.
The resulting $\widetilde q^e$ yields the signed coefficient and logit direction
\begin{equation}
 \boxed{
 d_{i,t}=\widetilde q^O_{i,t}-\widetilde q^S_{i,t},
 \quad u_{i,t}=d_{i,t}-\delta_{i,t}p_{i,t},\quad
 \delta_{i,t}=\mathbf1^\top d_{i,t}.
 }
 \label{eq:pvec_direction}
\end{equation}
The mass difference $\delta=M_O-M_S$ need not vanish, so $u$, rather
than $d$, is the exact logit gradient of $\sum_v d(v)\log p(v)$.
In Figure~\ref{fig:pvec_mechanism}, the paired contrast gives C a positive
coefficient and B a negative one. These propose which preferences to
strengthen or suppress; reward support determines whether to accept them.

\subsection{Credit--Risk Allocation}
\label{sec:cra}

CRA first qualifies each proposal using local and peer credit (CQ), then
allocates a shared response budget according to credit and directional
risk (RA).

\noindent\textbf{Local and Peer Credit.} With sampled-token one-hot vector $e_y$,
local reward alignment is
$b_{i,t}=A_i u_{i,t}^{\top}(e_{y_{i,t}}-p_{i,t})$.
The base variance $c^0_{i,t}=\operatorname{Var}_{p_{i,t}}(d_{i,t})$
moderates local acceptance. Because updates affect shared parameters,
we also test support from other questions. Partition prompts into
$K_{\mathrm{fold}}=4$ deterministic folds, placing all responses to a
question in the same fold $f(i)$. For a linear output head $z=Wa$,
let $a^0_{i,t}$ be the snapshot hidden input and $G_k$ the GRPO
reward-ascent gradient of $W$ contributed by fold $k$:
\begin{equation}
 I_{i,t,k}=
 \frac{\left\langle G_k,u_{i,t}(a^0_{i,t})^\top\right\rangle_F}
 {\max\!\left\{\|G_k\|_F\|a^0_{i,t}\|_2,\epsilon_I\right\}},
 \quad \mu_{i,t}^{(-f(i))}=\frac{1}{K_{\mathrm{fold}}-1}
 \sum_{k\ne f(i)}I_{i,t,k}.
 \label{eq:peer_influence}
\end{equation}
Excluding $f(i)$ removes the entire source-question fold, including all
sibling responses; all positions of $y_i$ share the three peer gradients
shown in Figure~\ref{fig:ravec_overview}. Averaging $I_{i,t,k}$ quantifies
cross-question output-head support for the proposed update
$u_{i,t}(a^0_{i,t})^\top$.

The numerator equals $u_{i,t}^{\top}G_k a^0_{i,t}$. Here
$G_k a^0_{i,t}$ is the logit change at the current hidden input for a
unit step along the fold gradient. A positive inner product means this
change has a component along the proposed visual correction.

With $[a]_+=\max(a,0)$ and $\alpha,\epsilon,\epsilon_I>0$,
qualified credit is
\begin{equation}
 \boxed{h_{i,t}=r_i m_{i,t}
 \frac{[b_{i,t}]_+}{[b_{i,t}]_++\alpha c^0_{i,t}+\epsilon}
 \left[\mu_{i,t}^{(-f(i))}\right]_+ .}
 \label{eq:cra_credit}
\end{equation}

\newpage
\noindent\begin{minipage}[t]{0.49\linewidth}
\vspace{0pt}\setlength{\parskip}{0.5\baselineskip}
\noindent
Both signals must be positive. Figure~\ref{fig:peer_consistency} shows
the distinction: blue positions pass both tests, whereas orange
positions agree locally but receive no credit because peer support is
nonpositive. Thus the peer test changes which corrections can receive
the response budget, rather than merely rescaling every local proposal.

\noindent\textbf{Directional Risk.} Equal-norm directions can have different
effects on uncertainty, as illustrated in Figure~\ref{fig:cra_mechanism}
(left). For predictive entropy $\mathcal H(p)=-\sum_v p_v\log p_v$,
the directional derivative is
\mbox{$\chi=\nabla_z\mathcal H(p)^\top u=-\operatorname{Cov}_p(\log p,u)$}.
\end{minipage}\hfill
\begin{minipage}[t]{0.48\linewidth}
  \vspace{10pt}\centering
  \captionsetup{type=figure,font=footnotesize,justification=raggedright,singlelinecheck=false,skip=4pt}
  \includegraphics[width=\linewidth]{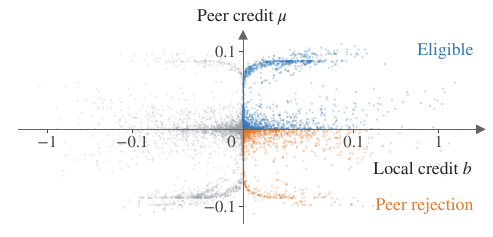}
  \caption{\textbf{CQ: selecting supported corrections.}
    Blue: $b>0,\mu>0$; orange: locally aligned but rejected by the peer test;
    gray: nonpositive local credit.}
  \label{fig:peer_consistency}
\end{minipage}\par
\noindent
Its sign distinguishes entropy expansion from contraction.
Writing $F=\operatorname{Var}_p(u)$ and $V=\operatorname{Var}_p(\log p)$,
\mbox{we define}
\begin{equation}
 \rho_{i,t}
 =\frac{[\min(\chi_{i,t},0)]^2}
 {(V_{i,t}+\epsilon_\rho)(F_{i,t}+\epsilon_\rho)},
 \quad C_{i,t}=\|u_{i,t}\|_2^2\bigl(1+\rho_{i,t}\bigr),
 \quad\epsilon_\rho>0.
 \label{eq:directional_cost}
\end{equation}
\par
\Needspace{3\baselineskip}
\noindent
Only locally entropy-contracting directions incur the additional cost.
Cauchy--Schwarz gives $0\le\rho\le1$, so the penalty at most doubles
the squared-norm cost. This bounded adjustment incorporates predictive
uncertainty into response-level allocation.

\Needspace{225pt}
\noindent\begin{minipage}[t]{0.49\linewidth}
\vspace{0pt}\setlength{\parskip}{0.5\baselineskip}
\noindent\textbf{Budget and Joint Allocation.} To coordinate correction strengths,
we set a response budget from the base credit
$b^{\mathrm{base}}_{i,t}=A_i(d_{i,t}(y_{i,t})-p_{i,t}^{\top}d_{i,t})$:
\begin{alignedEquation}{eq:cra_allocation}
 w^0_{i,t}&=r_i m_{i,t}
 \frac{[b^{\mathrm{base}}_{i,t}]_+}
 {[b^{\mathrm{base}}_{i,t}]_++\alpha c^0_{i,t}+\epsilon},\\
 B_i&=\sum_{t=1}^{T_i}C_{i,t}(w^0_{i,t})^2.
\end{alignedEquation}
Unlike $b$, the base credit uses the uncentered $d$; the two can differ
when retained masses differ. It sets the available energy before peer
selection, while $h$ determines its allocation.

Let $\mathcal A_i=\{t:r_i m_{i,t}=1,\ h_{i,t}>0,\ C_{i,t}>0\}$.
\end{minipage}\hfill
\begin{minipage}[t]{0.48\linewidth}
  \vspace{0pt}\centering
  \captionsetup{type=figure,font=footnotesize,justification=raggedright,singlelinecheck=false,skip=4pt}
  \includegraphics[width=\linewidth]{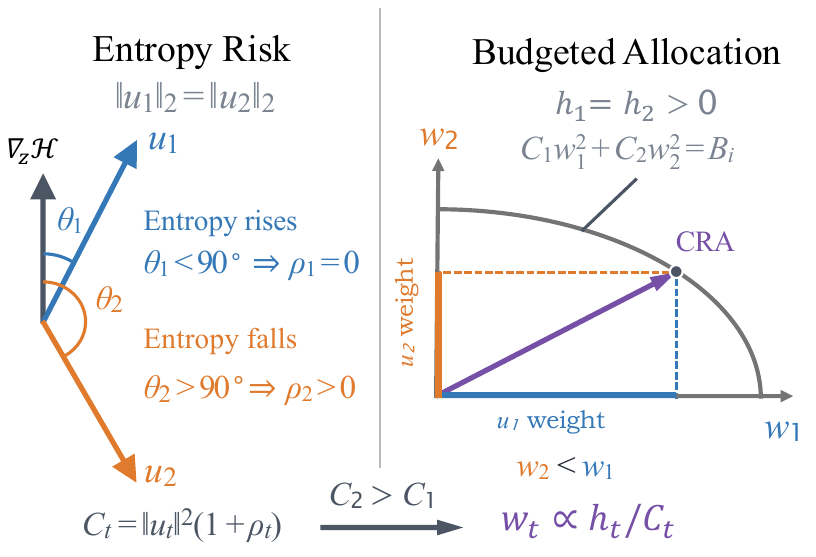}
  \caption{\textbf{RA: allocating the response budget.}
    At equal norm, entropy contraction raises directional cost (left).
    At equal credit, optimal allocation shifts toward the lower-cost
    direction (right).}
  \label{fig:cra_mechanism}
\end{minipage}\par
\noindent
We maximize $\sum_{t\in\mathcal A_i}h_{i,t}w_{i,t}$ over $w_{i,t}\ge0$
subject to $\sum_{t\in\mathcal A_i}C_{i,t}w_{i,t}^2\le B_i$.
With $D_i=\sum_{t\in\mathcal A_i}h_{i,t}^2/C_{i,t}$,
weighted Cauchy--Schwarz gives
\begin{equation}
 \boxed{w^{\mathrm{alloc}}_{i,t}=
 \begin{cases}
 \sqrt{B_i/D_i}\,h_{i,t}/C_{i,t},&
 t\in\mathcal A_i,\ D_i>0,\\
 0,&\text{otherwise}.
 \end{cases}}
 \label{eq:cra_normalization}
\end{equation}
The shared factor $\sqrt{B_i/D_i}$ couples all accepted positions.
Figure~\ref{fig:cra_mechanism} connects the two steps: entropy contraction
raises $C_2$ on the left, and at equal credit the solution on the right
gives $w_2/w_1=C_1/C_2<1$. No qualified position or a zero budget yields
zero allocation. Otherwise the budget is saturated; nonnegative
weights may exceed one.

\begin{algorithm}[!t]
\caption{RAVEC Training}
\label{alg:ravec}
\small
\color{black}
\hypersetup{citecolor=black,urlcolor=black}
\renewcommand{\algorithmiccomment}[1]{\algcomment{#1}}
\begin{algorithmic}[1]
\REQUIRE Actor $\pi_\theta$, EMA teacher $\pi_\phi$, data $\mathcal D$.
\STATE $\phi\gets\theta$; initialize AdamW~\citep{ref74}.
\REPEAT[Rollout-start cache]
\STATE $\theta_{\mathrm{old}}\gets\theta$;
$y_i\sim\pi_{\theta_{\mathrm{old}}}(\cdot\mid x_i,I_i^O)$;
$R_i\gets\operatorname{Verifier}(x_i,I_i^O,y_i)$.
\STATE $A_i\gets\frac{R_i-\bar R_{\mathcal G(i)}}{\sigma_{\mathcal G(i)}+10^{-6}}$;
cache $r,m,L,f,p,a^0,G_k$ from all valid tokens.
\STATE $d,u,h,C,w^0,w^{\mathrm{alloc}}\gets0$.
\FOR[Signed correction]{$(i,t)$ with $r_i m_{i,t}=1$}
\STATE Paired $q^e\to\mathcal S,M_e,\widetilde q^e$ at $\tau_v=2$ (Sec.~\ref{sec:pvec});
\alghl{algdirbg}{d\gets\widetilde q^O-\widetilde q^S};
$u\gets d-(\mathbf1^\top d)p$.
\STATE $b,b^{\mathrm{base}},c^0,\mu,\rho,C,w^0$ from Sec.~\ref{sec:cra};
\alghl{algcreditbg}{h\gets\frac{[b]_+}{[b]_++\alpha c^0+\epsilon}[\mu]_+}
\algcomment{Credit gate}
\ENDFOR
\STATE \textbf{for each response} $i$, form $\mathcal A_i,B_i,D_i$ (Sec.~\ref{sec:cra});
\textbf{if} $D_i>0$, for $t\in\mathcal A_i$:
\alghl{algallocbg}{w^{\mathrm{alloc}}_{i,t}\gets\sqrt{B_i/D_i}\,h_{i,t}/C_{i,t}}.
\STATE $\mathcal E(w)\gets\sum_i L_i^{-2}\sum_t m_{i,t}w_{i,t}^2\|u_{i,t}\|_2^2$;
$(E_{\mathrm{base}},E_{\mathrm{alloc}})\gets
(\mathcal E(w^0),\mathcal E(w^{\mathrm{alloc}}))$.
\STATE \textbf{if} $E_{\mathrm{alloc}}=0$ or nonfinite energies: \textbf{continue}.
\STATE $\zeta\gets\sqrt{E_{\mathrm{base}}/E_{\mathrm{alloc}}}$;
$(d,w^{\mathrm{used}})\gets\operatorname{sg}(d,\zeta w^{\mathrm{alloc}})$.
\FOR[Actor and EMA updates]{actor batch $\mathcal O$}
\STATE $q^{\mathrm{EMA}}_{i,t}\gets\operatorname{sg}
\bigl[\pi_\phi(\cdot\mid x_i,I_i^O,y_{i,<t})\bigr]$;
$\mathcal L_{\mathrm{total}}\gets\mathcal L_{\mathrm{GRPO}}
+\lambda_{\mathrm{vis}}\mathcal L_{\mathrm{vis}}+c_{\mathrm{ema}}\mathcal L_{\mathrm{ema}}$.
\STATE $\theta\gets\operatorname{AdamW}(\theta,
\operatorname{ClipNorm}(\nabla_\theta\mathcal L_{\mathrm{total}},1))$;
$\phi\gets\tau\phi+(1-\tau)\theta$.
\ENDFOR
\UNTIL{training budget exhausted}
\RETURN $\pi_\theta$
\end{algorithmic}
\end{algorithm}

\subsection{Optimization Objective and Training}
\label{sec:objective}

With $\zeta=\sqrt{E_{\mathrm{base}}/E_{\mathrm{alloc}}}$, the calibrated weights
$w^{\mathrm{used}}=\operatorname{sg}(\zeta w^{\mathrm{alloc}})$ match the base
weights' cached position-logit gradient energy, including response-length
normalization, and preserve relative allocations. The budget $B_i$ applies
before scaling. Algorithm~\ref{alg:ravec} skips all actor and EMA updates
for rollouts with nonfinite energies or $E_{\mathrm{alloc}}=0$.

For $s_{i,t}=\sum_v d_{i,t}(v)\log p_{\theta,i,t}(v)$ and
$L_i=\max(\sum_t m_{i,t},1)$, the visual loss averages response means:
\begin{equation}
 \mathcal L_{\mathrm{vis}}
 =-\frac{1}{N_{\mathrm{aux}}}
 \sum_{i\in\mathcal O}\frac{1}{L_i}\sum_{t=1}^{T_i}
 m_{i,t}w^{\mathrm{used}}_{i,t}s_{i,t}.
 \label{eq:visual_loss}
\end{equation}
Here $N_{\mathrm{aux}}=|\mathcal O|$ counts all responses in the optimizer batch,
including those with zero visual loss. A separate reference loss $\mathcal L_{\mathrm{ema}}$
averages $D_{\mathrm{KL}}(q^{\mathrm{EMA}}_{i,t}\|p_{\theta,i,t})$ over
the same valid tokens as GRPO, independently of visual acceptance.
The detached, full-vocabulary EMA reference~\citep{ema_pg2026} is refreshed
at temperature one before each actor step. We optimize
\begin{equation}
 \mathcal L_{\mathrm{total}}
 =\mathcal L_{\mathrm{GRPO}}
 +\lambda_{\mathrm{vis}}\mathcal L_{\mathrm{vis}}
 +c_{\mathrm{ema}}\mathcal L_{\mathrm{ema}}.
 \label{eq:total}
\end{equation}
PVEC coefficients and CRA weights are cached per rollout with $\operatorname{sg}$ (stop-gradient); actor probabilities remain differentiable in the visual loss.
After each actor step, the teacher is updated by
$\phi\leftarrow\tau\phi+(1-\tau)\theta$.

\begin{table}[!t]
\centering
\caption{\textbf{Main results.}
Avg@8 accuracy (\%) on ten benchmarks. Backbones are Qwen2.5-VL-Instruct before RL.
Bold/underlined scores are best/second best per backbone.
MV-V/CVQA/MARS denote MathVerse-V/C-VQA-Real/MARS-Bench.}
\label{tab:main_results}
\begingroup
\hypersetup{citecolor=black}
\footnotesize
\newcommand{\methodsource}[3]{\shortstack[l]{%
  #1 \textcolor{black!65}{\scriptsize(#3)}\\[-0.4ex]
  \scriptsize\citep{#2}}}
\setlength{\tabcolsep}{1.5pt}
\renewcommand{\arraystretch}{1.12}
\begin{tabular*}{\linewidth}{@{\extracolsep{\fill}}l*{11}{c}@{}}
\toprule
Method & Geo3K & MVista & We-Math & MVerse & LVista & MMMU-Pro & MV-V & CVQA & MARS & POPE & Avg \\
\midrule
\shadebackbonerow
\textbf{Qwen2.5-VL-3B} & 18.72 & 46.49 & 24.09 & 28.56 & 28.71 & 17.90 & 30.55 & 37.08 & 28.48 & 33.52 & 29.41 \\
\midrule
\hspace{0.5em}\methodsource{GRPO}{ref15}{arXiv'24} & 28.72 & 59.34 & 58.90 & 55.25 & 38.14 & 25.66 & 52.26 & 63.76 & 48.35 & 87.09 & 51.75 \\
\hspace{0.5em}\methodsource{DAPO}{ref17}{NeurIPS'25} & \underline{31.20} & \underline{60.89} & 59.95 & 56.25 & 40.69 & \textbf{28.42} & 53.09 & \underline{65.27} & \underline{50.26} & \textbf{88.37} & \underline{53.44} \\
\hspace{0.5em}\methodsource{PAPO}{ref56}{ICLR'26} & 30.95 & \textbf{61.38} & 60.09 & 57.14 & 38.67 & 27.11 & \underline{53.95} & 63.78 & 49.72 & 86.49 & 52.93 \\
\hspace{0.5em}\methodsource{CFPO}{ref80}{ICML'26} & 30.87 & 59.71 & \underline{60.68} & \underline{57.36} & \underline{40.82} & 28.33 & 53.14 & 64.21 & 49.39 & 87.92 & 53.24 \\
\hspace{0.5em}\textbf{RAVEC (ours)} & \textbf{33.28} & 59.78 & \textbf{61.68} & \textbf{57.43} & \textbf{41.67} & \underline{28.38} & \textbf{54.44} & \textbf{66.54} & \textbf{53.05} & \underline{87.98} & \textbf{54.42} \\
\midrule
\shadebackbonerow
\textbf{Qwen2.5-VL-7B} & 35.40 & 63.50 & 33.00 & 31.47 & 39.70 & 24.36 & 34.58 & 58.83 & 42.74 & 79.51 & 44.31 \\
\midrule
\hspace{0.5em}\methodsource{GRPO}{ref15}{arXiv'24} & 40.18 & 65.48 & 68.12 & 66.51 & 45.62 & 35.17 & 61.71 & 70.37 & 55.21 & \underline{87.72} & 59.61 \\
\hspace{0.5em}\methodsource{DAPO}{ref17}{NeurIPS'25} & 35.92 & 61.91 & 58.51 & 55.64 & 37.05 & 29.02 & 51.04 & 70.07 & 55.66 & 87.21 & 54.20 \\
\hspace{0.5em}\methodsource{PAPO}{ref56}{ICLR'26} & 40.25 & \textbf{69.53} & 66.79 & \textbf{68.43} & 46.07 & \underline{36.63} & \underline{64.97} & 69.76 & 54.59 & 87.70 & \underline{60.47} \\
\hspace{0.5em}\methodsource{TPAE}{ref81}{ACM MM'26} & \textbf{46.73} & 64.20 & \underline{69.07} & 45.50 & \textbf{47.20} & \textbf{38.27} & 64.37 & \textbf{72.29} & \textbf{58.10} & 87.37 & 59.31 \\
\hspace{0.5em}\textbf{RAVEC (ours)} & \underline{43.20} & \underline{66.44} & \textbf{70.73} & \underline{67.53} & \underline{46.28} & 36.09 & \textbf{65.03} & \underline{70.69} & \underline{55.99} & \textbf{87.99} & \textbf{61.00} \\
\bottomrule
\end{tabular*}
\endgroup
\end{table}

\section{Experiments}
\label{sec:experiments}

\subsection{Experimental Setup}
We evaluate the 3B and 7B Qwen2.5-VL-Instruct~\citep{qwen25vl2025} models
under GRPO on ten benchmarks. Main training uses 38,870
ViRL39K~\citep{virl39k2025} questions.

Following the evaluation protocol of PAPO~\citep{ref56}, we use
exact-match scoring with a sampling temperature of 1.0 and report
Avg@8, the mean accuracy over eight generations per question.
Evaluations use actor weights, and the table's Avg is the equal-weight
mean across the ten evaluation benchmarks.

\paragraph{Default Configuration.}
The loss weights in Eq.~\eqref{eq:total} are
$\lambda_{\mathrm{vis}}=0.5$ and $c_{\mathrm{ema}}=0.1$.
Teacher and parameter comparisons use a fixed evaluation subset, with
default base-cost coefficient $\alpha=0.0025$,
$\epsilon=2.5\!\times\!10^{-5}$, per-endpoint Top-$K$ budget $K=128$,
EMA decay $\tau=0.95$, learning rate $10^{-6}$, and top-$p=1$.
Each ablation varies the indicated setting.

\paragraph{Controlled Training Comparison.}
For Section~\ref{sec:further_analysis}, we compare GRPO \citep{ref15},
PAPO \citep{ref56}, VGPO \citep{ref79}, TPAE \citep{ref81}, CFPO \citep{ref80},
and RAVEC using Qwen2.5-VL-3B-Instruct,
the same 1,024 training questions, 200 training steps, and seed 1.
We evaluate Avg@8 every 20 steps on a fixed validation set comprising
20 questions from each of eight benchmarks.

\subsection{Main Results}

Table~\ref{tab:main_results} shows that RAVEC achieves the highest average
across ten benchmarks at both model scales.
At 3B, RAVEC averages 54.42, outperforming the strongest baseline,
DAPO~\citep{ref17}, by 0.98 percentage points and GRPO by 2.67 percentage
points.
At 7B, RAVEC reaches 61.00, improving on PAPO by 0.53 points and GRPO
by 1.39 points.

The improvements over GRPO cover all ten tasks at each scale.
At 3B, the largest gains are 4.56 points on Geometry3K and 4.70 on
MARS-Bench, with further gains on We-Math and LogicVista.
At 7B, improvements of 2.61 points on We-Math and 3.32 on MathVerse-V
show that the average gain extends across multiple reasoning tasks.

At 3B, RAVEC ranks first on seven of ten benchmarks. Its margins over the next-best method are 2.79 percentage points on MARS-Bench, 2.08 on Geometry3K, 1.27 on C-VQA-Real, 1.00 on We-Math, 0.85 on LogicVista, 0.49 on MathVerse-V, and 0.07 on MathVerse. At 7B, RAVEC ranks first on We-Math, POPE, and MathVerse-V, with margins of 1.66, 0.27, and 0.06 points, respectively.

\begin{table}[!t]
\centering
\caption{\textbf{Component, teacher, and parameter ablations.}
Results for the 3B model in percent.
Panel a reports full test scores for component ablations and fixed subset
scores for teacher and cost comparisons. Avg is the unweighted mean across
eight benchmarks. Panels b and c report question-weighted final accuracy
and checkpoint means with sample standard deviations on the same subset.
EA, CQ, and RA denote evidence alignment, credit qualification, and risk-aware allocation, respectively.}
\label{tab:ablations}
\label{tab:analysis}
\label{tab:sensitivity}
\label{tab:validation}
\small
\setlength{\tabcolsep}{3pt}
\renewcommand{\arraystretch}{1.12}
\noindent\textbf{(a) Component ablations}\par\vspace{3pt}
\begin{tabular*}{\linewidth}{@{\extracolsep{\fill}}l*{9}{c}@{}}
\toprule
Setting & Geo3K & MVista & We-Math & MVerse & LVista & Count & MMMU-Pro & MV-V & Avg \\
\midrule
\textbf{RAVEC (ours)} & 33.28 & 59.78 & 61.68 & 57.43 & 41.67 & 75.63 & 28.38 & 54.44 & \textbf{51.54} \\
\hspace{0.5em}w/o EA & 32.05 & 58.54 & 60.37 & 56.49 & 41.83 & 73.85 & 28.73 & 53.78 & 50.71 \\
\hspace{0.5em}w/o CQ & 32.92 & 57.46 & 60.52 & 56.66 & 41.14 & 71.19 & 28.67 & 54.54 & 50.39 \\
\hspace{0.5em}w/o RA & 33.17 & 59.56 & 60.24 & 57.56 & 39.46 & 60.25 & 29.03 & 54.81 & 49.26 \\
\midrule
\multicolumn{10}{@{}l}{\textit{Teacher source and EMA decay}} \\
\midrule
Fixed teacher & 22.29 & 51.17 & 48.93 & 49.46 & 33.90 & 46.69 & 23.25 & 41.44 & 39.64 \\
EMA 0.90 & 26.65 & 52.19 & 53.99 & 52.87 & 34.53 & 49.88 & 23.25 & 45.04 & \textbf{42.30} \\
EMA 0.95 & 24.66 & 53.26 & 50.44 & 49.66 & 36.72 & 49.06 & 22.52 & 42.51 & 41.10 \\
EMA 0.99 & 24.85 & 52.38 & 49.90 & 50.83 & 32.73 & 46.44 & 23.30 & 43.04 & 40.44 \\
\midrule
\multicolumn{10}{@{}l}{\textit{Base cost coefficient}} \\
\midrule
$\alpha=0.00125$ & 25.44 & 53.50 & 51.46 & 53.06 & 36.62 & 47.06 & 25.24 & 44.16 & 42.07 \\
$\alpha=0.00250$ & 24.66 & 53.26 & 50.44 & 49.66 & 36.72 & 49.06 & 22.52 & 42.51 & 41.10 \\
$\alpha=0.00500$ & 24.85 & 52.82 & 51.80 & 54.62 & 35.26 & 54.50 & 24.27 & 44.46 & \textbf{42.82} \\
\bottomrule
\end{tabular*}

\par\medskip
\begin{minipage}[t]{0.48\linewidth}
\noindent\textbf{(b) Visual-loss coefficient}\par\vspace{3pt}
\begin{tabular*}{\linewidth}{@{\extracolsep{\fill}}lrr@{}}
\toprule
$\lambda_{\mathrm{vis}}$ & Avg@8 & Mean $\pm$ SD \\
\midrule
0.25 & 40.87 & 38.33$\mkern2mu\pm\mkern2mu$2.42 \\
0.50 & 40.87 & 38.54$\mkern2mu\pm\mkern2mu$2.82 \\
0.75 & 44.09 & 40.27$\mkern2mu\pm\mkern2mu$3.86 \\
1.00 & 45.40 & \textbf{41.32}$\mkern2mu\pm\mkern2mu$4.16 \\
\bottomrule
\end{tabular*}
\end{minipage}\hfill
\begin{minipage}[t]{0.48\linewidth}
\noindent\textbf{(c) Per-endpoint Top-$K$}\par\vspace{3pt}
\begin{tabular*}{\linewidth}{@{\extracolsep{\fill}}lrr@{}}
\toprule
$K$ & Avg@8 & Mean $\pm$ SD \\
\midrule
64 & 41.74 & 38.93$\mkern2mu\pm\mkern2mu$2.96 \\
128 & 40.87 & 38.54$\mkern2mu\pm\mkern2mu$2.82 \\
256 & 41.73 & \textbf{39.03}$\mkern2mu\pm\mkern2mu$2.54 \\
\bottomrule
\end{tabular*}
\end{minipage}
\end{table}

\subsection{Component Analysis}
\label{sec:parameter_comparisons}

Table~\ref{tab:ablations}, panel a, examines evidence alignment,
credit qualification, and risk-aware allocation.
All variants retain the EMA KL stabilization term~\citep{ema_pg2026}. Removing EA permutes suppression boxes across matched samples,
preserving the original image and teacher replay.
Average accuracy drops 0.83 points, from 51.54 to 50.71,
with losses on six tasks.
Counting drops by 1.78 points, We-Math by 1.31, MathVista by 1.24,
and Geometry3K by 1.23. These results support aligning suppression
with the evidence region to obtain useful paired visual corrections.

Removing CQ disables local and peer qualification.
Average accuracy drops 1.15 points, including losses of 4.44 on
Counting and 2.32 on MathVista.
This supports screening corrections for reward agreement before
visual guidance.

Replacing RA removes directional risk and uniformly weights active
tokens under the same response energy budget.
The average falls by 2.28 points, mainly due to a 15.38-point loss on
Counting, with additional losses on LogicVista and We-Math.
Counting benefits most from the complete design in all three comparisons.
These ablations support the combined design: CQ couples local and peer
credit; RA couples directional risk and credit-based weighting. RAVEC
outperforms all ablations on Geometry3K, MathVista, We-Math, and Counting.

\subsection{Teacher and Parameter Analysis}
Table~\ref{tab:analysis} compares teachers and parameters on a fixed
evaluation subset. The default EMA teacher raises the average from
39.64 with a fixed teacher to 41.10, while decay 0.90 reaches 42.30.
These short runs favor a teacher that adapts during training.
Teacher choice jointly affects visual guidance and reference regularization
in this comparison. Increasing the visual loss coefficient from 0.25 to 1.00 raises final
validation accuracy from 40.87 to 45.40.
The checkpoint mean increases from 38.33 to 41.32, while its sample
standard deviation rises from 2.42 to 4.16.
Checkpoint summaries capture the resulting learning dynamics.

The best tested base cost coefficient is 0.005, with an average of
42.82. Both 0.005 and 0.00125 offer promising settings for tuning
around the default 0.0025.
Support budgets of 64 and 256 yield similar final scores of 41.74 and
41.73, suggesting limited sensitivity over the tested range.
These subset comparisons characterize teacher and parameter behavior,
complementing the full test results in Table~\ref{tab:main_results}.

\begin{figure}[!htbp]
    \centering
    \includegraphics[width=\textwidth]{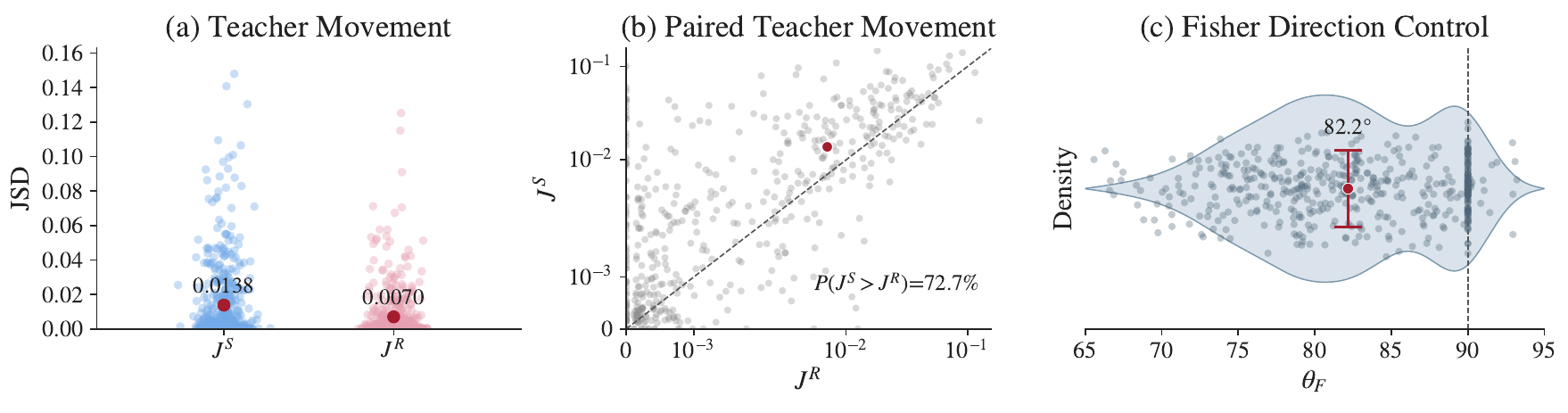}
    \caption{\textbf{Evidence region control.}
    Evidence and random suppression produce the prompt means $J^S$ and
    $J^R$ of tokenwise Jensen-Shannon divergence from the original view.
    Panels a and b show marginal and paired comparisons.
    Panel c shows the Fisher angle between the evidence and random
    suppression correction directions.}
    \label{fig:pvec_direction_motivation}
\end{figure}

\subsection{Visual Correction Diagnostics}
Figure~\ref{fig:pvec_direction_motivation} tests whether the suppressed
region changes teacher guidance.
The random control matches area with negligible evidence overlap;
teacher replay uses the same prefixes.
Evidence suppression produces a mean Jensen-Shannon divergence~\citep{ref73} of
0.0138, compared with 0.0070 for the control, and a larger shift on
72.7 percent of prompts. The mean Fisher angle~\citep{ref71} of 82.2 degrees also indicates distinct correction
directions. Thus, matched suppression area does not make the
interventions equivalent: they differ in both teacher movement and
correction direction.
This offline diagnostic characterizes the proposed directions,
while the EA ablation evaluates their effect on accuracy.

\subsection{Further Analysis}
\label{sec:further_analysis}

\begin{figure}[!htb]
    \centering
    \includegraphics[width=\linewidth]{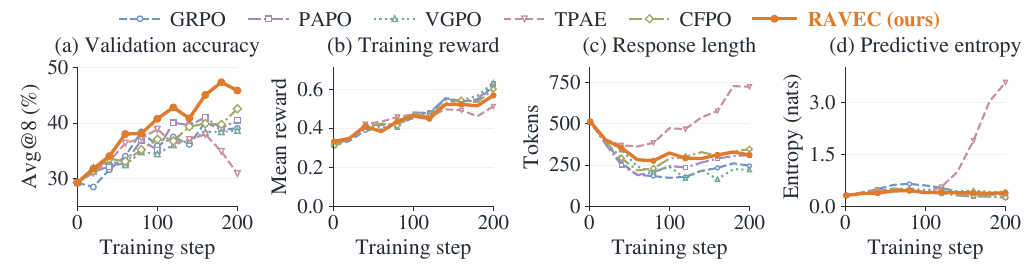}\par
    \caption{\textbf{Training dynamics of the 3B model.}
    Validation accuracy, training reward, response length, and predictive
    entropy under the shared setup in Section~\ref{sec:experiments}.
    Training curves show trailing means over five steps at intervals of
    20 steps. Validation points are unsmoothed.}
    \label{fig:ravec_training_dynamics}
    \medskip
    \nextfloat
    \includegraphics[width=\linewidth]{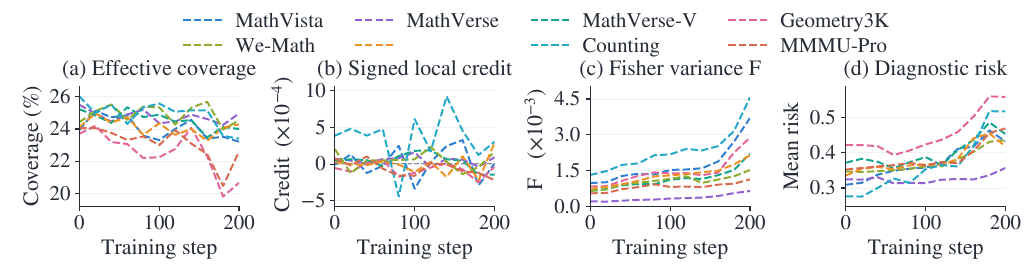}\par
    \caption{\textbf{Correction diagnostics across tasks.}
    Coverage uses $w^{\mathrm{used}}>0.001$. Panels b--d show pre-gate credit,
    Fisher direction variance $F$, and diagnostic risk $\rho_{\mathrm{diag}}$
    over eligible tokens; $\rho_{\mathrm{diag}}$ may differ from training risk in Eq.~\eqref{eq:directional_cost}.}
    \label{fig:ravec_task_mechanism}
    \medskip
    \nextfloat
    \includegraphics[width=\linewidth]{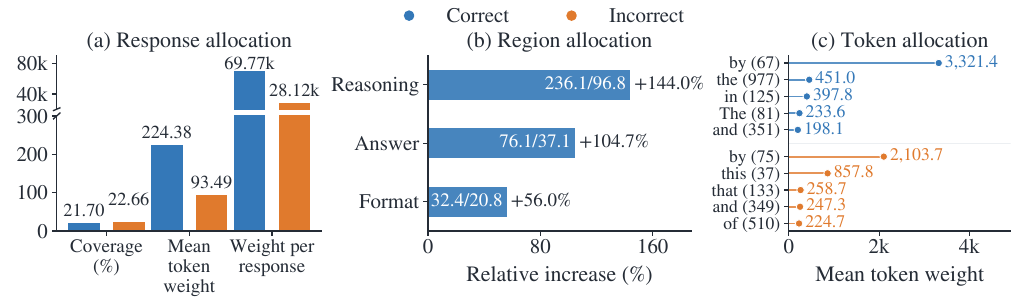}\par
    \caption{\textbf{Allocation in 100 matched response pairs.}
    Panel a compares coverage above 0.001, mean token weight, and response
    weight sums on a broken axis, with k denoting thousands.
    Panel b shows regional increases with correct then incorrect means.
    Panel c ranks the five alphabetic token types with the highest mean weight
    per group and gives counts.}
    \label{fig:ravec_weight_allocation}
\end{figure}

\paragraph{Training Dynamics.}
Figure~\ref{fig:ravec_training_dynamics} reports 45.86 percent final
validation accuracy for RAVEC, 3.28 points above
CFPO, the best final baseline.
GRPO and VGPO have higher training rewards but lower validation accuracy.
Late in training, TPAE responses lengthen and entropy rises as validation
accuracy declines.
RAVEC maintains comparatively steady entropy, and response length
settles after the initial decrease.
RAVEC peaks at 47.34 percent, and its final accuracy exceeds every
baseline's highest observed score. These trajectories summarize validation performance alongside training
rewards, response lengths, and predictive entropy.

\paragraph{Correction Dynamics.}
Figure~\ref{fig:ravec_task_mechanism} shows effective coverage around
20 to 26 percent, with other diagnostics varying across tasks.
Signed local credit measures agreement before the gate selects
positions with positive local and peer support for nonnegative allocation.
Fisher direction variance $F$ generally rises late in training, with
Counting showing the largest final value.
Geometry3K combines the
lowest final coverage with the highest diagnostic entropy risk.
MathVerse ends with the lowest variance and risk, despite coverage
comparable to several other tasks. These diagnostics characterize the
correction directions, not the complete allocation cost $C$. The late
rise in variance and risk without a comparable increase in coverage
indicates changing correction geometry rather than simply supervising
more positions.

\paragraph{Weight Allocation.}
Figure~\ref{fig:ravec_weight_allocation} compares 100 correct and incorrect
response pairs matched by question.
All summaries use applied weights, including zeros. The ``Answer''
region follows the reasoning boundary; ``Format'' includes tags and
other boundary tokens. Ranked token types occur at least 20 times in
each group.
Correct responses receive 2.40 times the mean token weight,
despite slightly lower coverage of 21.70 percent versus 22.66 percent.
Their larger average weight therefore reflects correction strength
rather than broader coverage. The mean weight sum per response is also
higher for correct answers, at 69.77k versus 28.12k, indicating greater
total allocation per response. Supplementary Table~\ref{tab:app_pair_summary} shows
that mean lengths differ by 3.4 percent, versus 148.1 percent for weight sums.
The relative increase is largest in reasoning tokens at 144.0 percent,
versus 104.7 percent for answers and 56.0 percent for formatting.
This pattern associates stronger allocation with
intermediate reasoning within the matched responses.
The token ``by'' leads both rankings, while other high-weight types
differ, highlighting variation in allocation within responses.

Correct responses receive larger weights with slightly lower coverage,
and reasoning shows the largest relative increase. Overall weight and coverage differences have paired bootstrap intervals
spanning zero.
One token accounts for 64.38 percent of the correct-response weight. This
concentration makes pooled averages sensitive to individual tokens, so
question-level comparisons remain important. These associations do not
establish causality.

\section{Conclusion}
\label{sec:conclusion}

We introduced RAVEC for reward-aligned visual correction.
PVEC proposes signed directions from paired teacher predictions; CRA
qualifies and jointly weights them using local and cross-question
reward support, a response budget, and directional entropy cost.
Experiments across ten benchmarks and two model scales show the highest
average among baselines, with component ablations supporting the design.
Correction operates only during training. Future work can extend the
framework through automatic region selection and broader model coverage.

\section*{LLM Usage Statement}

We used OpenAI ChatGPT and Codex to assist with manuscript revision and
language polishing, figure preparation and revision, and code implementation,
debugging, and maintenance. The authors directed this assistance and retain
full responsibility for the manuscript, figures, code, experimental results,
and conclusions.

\FloatBarrier
\begingroup
\interlinepenalty=10000
\hypersetup{urlcolor=black}
\newif\ifravecprintedbib
\IfFileExists{main.bib}{%
  \IfFileExists{ravec-abbrvnat.bst}{%
    \bibliographystyle{ravec-abbrvnat}%
    \AddToHookNext{cmd/bibitem/before}{\global\ravecprintedbibtrue}%
    \bibliography{main}%
  }{}%
}{}%
\ifravecprintedbib\else

\fi
\endgroup

\clearpage
\appendix
\makeatletter
\setlength{\@fptop}{0pt}
\setlength{\@fpsep}{12pt}
\makeatother
\providecommand{\theHalgorithm}{}
\renewcommand{\theHalgorithm}{appendix.\arabic{algorithm}}
\renewcommand{\theHfigure}{appendix.\arabic{figure}}
\renewcommand{\theHtable}{appendix.\arabic{table}}
\renewcommand{\theHequation}{appendix.\arabic{equation}}
\setcounter{figure}{0}
\renewcommand{\thefigure}{S\arabic{figure}}
\setcounter{table}{0}
\renewcommand{\thetable}{S\arabic{table}}
\setcounter{equation}{0}
\renewcommand{\theequation}{S\arabic{equation}}
\begingroup
\centering
\Large
\textbf{\papertitle}\\[0.2em]
Supplementary Material\par
\vspace{0.5em}
\endgroup

\section{Notation and Conventions}
\label{sec:app_notation}

\begingroup
\small
\setlength{\tabcolsep}{4pt}
\setlength{\LTpre}{6pt}
\setlength{\LTpost}{8pt}
\setlength{\LTcapwidth}{\textwidth}
\renewcommand{\arraystretch}{1.05}
\begin{longtable}{@{}L{.29\linewidth}L{.47\linewidth}L{.20\linewidth}@{}}
\caption{\textbf{Symbol and equation index.} Definitions cover the method,
loss reductions, and offline analyses. Vocabulary vectors are
columns; indices and reused letters take their local meanings.}
\label{tab:app_notation}\\
\toprule
Symbols & Object and role & Location \\
\midrule
\endfirsthead
\toprule
Symbols & Object and role & Location \\
\midrule
\endhead
\bottomrule
\endfoot
\multicolumn{3}{@{}l}{\textbf{Data, policies, and conventions}}\\
$x_i,I_i^O,I_i^S,\mathcal R_i$ & Question, original/suppressed images, and evidence region. & Sec.~\ref{sec:method} \\
$y_i,y_{i,<t},m_{i,t},T_i,L_i$ & Response, prefix, valid mask, padded and normalization lengths. & \eqref{eq:app_visual_loss} \\
$r_i,A_i,R_i,\mathcal G(i),n$ & Evidence flag; advantage/reward; sibling group and its size. & Sec.~\ref{sec:method} \\
$\theta,\theta_{\rm old},\phi,\tau$ & Actor, sampling policy, EMA teacher, and EMA decay. & Alg.~(\ref{alg:app_training}) \\
$z,p,q^O,q^S,q^{\rm EMA}$ & Actor logits/probabilities; paired teachers and EMA reference. & \eqref{eq:app_softening}, \eqref{eq:app_ema_average} \\
\addlinespace[2pt]
\multicolumn{3}{@{}l}{\textbf{Visual direction and local/peer credit}}\\
$\mathcal V,\mathcal S,K,M_e$ & Vocabulary/support sets, candidates per endpoint, retained mass. & \eqref{eq:app_support} \\
$\widetilde q^e,\tau_v,d^{\rm raw},d^{\rm soft}$ & Mass-preserving softened endpoint, temperature, raw/soft contrasts. & \eqref{eq:app_softening} \\
$U_p(a),N_p(a),\gamma$ & Logit gradient map, squared norm, and diagnostic scale. & \eqref{eq:app_norm}--\eqref{eq:app_norm_matching} \\
$d,\delta,u^0,\widehat u(\theta)$ & Cached coefficient and sum, initial and live logit gradients. & \eqref{eq:app_signed_score}--\eqref{eq:app_signed_gradient} \\
$e_y,v=e_y-p,b^{\rm base},c^0$ & Token basis, score gradient; base credit and $\operatorname{Var}_p(d)$. & \eqref{eq:app_base_credit} \\
$b,h,\alpha,\epsilon$ & Actual local credit, qualified credit, and gate constants. & \eqref{eq:app_local_credit}, \eqref{eq:app_credit_budget} \\
$W,a^0,G_k,f(i)$ & Output head, initial hidden input, fold ascent, question fold. & \eqref{eq:fold_gradient} \\
$I_{i,t,k},\mu^{(-f(i))}_{i,t}$ & Normalized head agreement and mean other-fold credit. & \eqref{eq:app_peer} \\
\addlinespace[2pt]
\multicolumn{3}{@{}l}{\textbf{Entropy risk, allocation, and calibration}}\\
$H,\chi,F,V$ & Entropy, directional derivative, $\operatorname{Var}_p(u)$, $\operatorname{Var}_p(\log p)$. & \eqref{eq:app_entropy_gradient}--\eqref{eq:app_entropy_moments} \\
$\rho,\epsilon_\rho,C$ & Stabilized entropy risk and cost $\|u^0\|_2^2(1+\rho)$. & \eqref{eq:app_risk} \\
$w^0,B_i,\mathcal A_i$ & Base weight, response budget, positive-credit active set. & \eqref{eq:app_credit_budget}--\eqref{eq:app_allocation_problem} \\
$D_i,w^{\rm alloc}$ & Allocation normalizer $\sum_{t\in\mathcal A_i}h_t^2/C_t$ and optimal weights. & \eqref{eq:app_allocation_bound}--\eqref{eq:app_allocation_solution} \\
$\mathcal E(a),E_{\rm base},E_{\rm alloc}$ & Initial position-logit energy; base and allocated totals. & \eqref{eq:calibration_energy} \\
$\zeta,w^{\rm used}$ & Batch calibration and final correction weight. & \eqref{eq:calibration_scale} \\
\addlinespace[2pt]
\multicolumn{3}{@{}l}{\textbf{Losses and reduction sets}}\\
$\mathcal O,\mathcal B,\mathfrak B(\mathcal O),N_{\rm aux},N_{\rm tok}$ & Update and microbatch sets, global response and token counts. & \eqref{eq:app_visual_loss}--\eqref{eq:app_pg_average} \\
$r^{\rm IS},\bar r^{\rm IS},\epsilon_{\rm clip},\ell^{\rm PG}$ & Likelihood ratios, symmetric clipping, and token loss. & \eqref{eq:app_pg_average}--\eqref{eq:app_pg_scalar} \\
$\mathcal L_{\rm vis},\operatorname{sg}(\cdot)$ & Response-mean visual loss; constant in one differentiation. & \eqref{eq:app_visual_loss} \\
$\mathcal L_{\rm ema},N_{\rm tok}$ & Forward KL averaged over all valid tokens; token count. & \eqref{eq:app_ema_average}--\eqref{eq:app_kl_gradient} \\
$\lambda_{\rm vis},c_{\rm ema},R$ & Loss coefficients and worker count. & \eqref{eq:total}; Alg.~(\ref{alg:app_training}) \\
\addlinespace[2pt]
\multicolumn{3}{@{}l}{\textbf{Diagnostic summaries}}\\
$\mathcal Q,\mathcal I(x),N_q,J_i^e,J^e$ & Question/response sets and count; response/question JS means. & \eqref{eq:app_region_js} \\
$q^R,\widehat q^e,\mathcal F(p),\theta_F$ & Random-view/softened teachers; Fisher metric and angle. & \eqref{eq:app_region_angle} \\
$T,q_T^e,\bar u_{K,T},u_{K,T}$ & Temperature, full-softmax teacher, pre/post finite-support gradients. & \eqref{eq:app_support_errors} \\
$\bar u_{\infty,T},u_{\infty,T},R_{\rm pre},R_{\rm post}$ & Full-vocabulary counterparts; pre/post relative errors. & \eqref{eq:app_support_errors} \\
$\mathcal E_{s,d,c},\langle x\rangle_{s,d,c}$ & Eligible positions by step, task, and outcome; pooled mean. & \eqref{eq:app_token_pool} \\
$\overline w_i,S_i,c_i$ & Response mean weight, weight sum, and effective coverage. & \eqref{eq:app_pair_weights}--\eqref{eq:app_pair_coverage} \\
\addlinespace[2pt]
\multicolumn{3}{@{}l}{\textbf{Fixed-prefix and virtual-head probes}}\\
$\mathcal C,z_{\mathcal C},v,\bar v,s(v)$ & Four options/logits; direction, centered form, and RMS. & \eqref{eq:app_choice_rms} \\
$\eta,g,\Delta\ell$ & Perturbation RMS, gold option, and its log-odds change. & \eqref{eq:app_gold_logodds} \\
$\xi,\mathcal I_i^{\rm loc},\mathcal I_i^{\rm peer},\mathcal I_i^{\rm top}$ & Continuous local gate; local, peer, count-matched candidates. & Sec.~\ref{sec:app_probes} \\
$E_{i,t},h^{(j)},C^{(j)},B_i^{\rm phys},w^{(j)}$ & Direction norm squared; four scores/costs, physical budget, weights. & Sec.~\ref{sec:app_probes} \\
$\Delta W_{\mathcal C}^{(j)},a_{i,t},a_q,\Delta z_q^{(j)}$ & Virtual head change, fixed hidden states, induced option logits. & \eqref{eq:app_virtual_head} \\
$\mathcal Q_{\rm cal},\sigma_j,z'_{q,\mathcal C}$ & Calibration questions, pooled RMS, perturbed option logits. & \eqref{eq:app_virtual_head} \\
\end{longtable}
\endgroup

Several sampled responses $i$ may share the multimodal question $x_i$.
A response has $T_i$ padded positions, with $m_{i,t}\in\{0,1\}$ marking
valid generated tokens. We define $L_i=\max\{\sum_t m_{i,t},1\}$;
padding never contributes to a numerator. Indices $i,t,v,k$ denote
responses, positions, vocabulary entries, and question folds. The
evidence indicator $r_i$ is one only for a valid localized region.
Zero-advantage responses receive no visual correction but remain in
the policy and reference denominators.

\paragraph{Differentiation and Update Timing.}
Algorithm~(\ref{alg:ravec}) constructs all PVEC and CRA quantities once
at rollout start. The teacher endpoints, support, coefficient $d$,
snapshot direction $u^0$, local and peer credits, risk, costs, budgets,
allocation, and calibration factor remain fixed throughout optimization
of that rollout. The operator $\operatorname{sg}$ removes their gradient
paths, and the rollout cache preserves them across Actor steps.

Within the correction definitions, $p,z,u$ refer to the initial
snapshots $p^0,z^0,u^0$. The superscript in $c^0$ and $w^0$ denotes the
base variance and base weight, which are also cached.

Current probabilities $p_\theta$ are used to evaluate the losses.
Their visual loss gradient can change even though $d$ and the allocation
remain fixed, as derived in Equation~\eqref{eq:app_signed_gradient}.

This change does not trigger new credit, risk, or weight calculations.

The original view EMA reference is recomputed before each Actor step,
held constant during that step, and followed by the teacher update.

\section{Deriving Visual Correction and Allocation}
\label{sec:app_derivations}
\label{sec:implementation}

\subsection{Finite Support and Mass-Preserving Softening}
At a fixed question and response prefix, let $q^O,q^S$ be normalized
teacher distributions. The support is the unique union of each endpoint's
$K$ largest-probability entries and the sampled token $y$:
\begin{equation}
 \mathcal S=\operatorname{Top}_K(q^O)\cup
 \operatorname{Top}_K(q^S)\cup\{y\},\quad |\mathcal S|\le 2K+1.
 \label{eq:app_support}
\end{equation}
Thus $K$ counts entries per endpoint. For $e\in\{O,S\}$, write
$M_e=\sum_{v\in\mathcal S}q^e_v$ and use temperature softening~\citep{ref33}:
\begin{equation}
 \widetilde q^e_v=\mathbf1[v\in\mathcal S]M_e
 \frac{(q^e_v)^{1/\tau_v}}
 {\sum_{j\in\mathcal S}(q^e_j)^{1/\tau_v}},\qquad \tau_v=2.
 \label{eq:app_softening}
\end{equation}
For finite teacher logits the denominator is positive. Summing over
$\mathcal S$ cancels it, giving $\sum_v\widetilde q^e_v=M_e$.
Consequently the softened contrast $d^{\rm soft}=\widetilde q^O-
\widetilde q^S$ has sum $M_O-M_S$, which need not vanish. No probability
mass outside the union is redistributed into it.

\paragraph{Training Direction and Diagnostic Scale.}
The training coefficient is $d=d^{\rm soft}$, exactly as in
Equation~\eqref{eq:pvec_direction}, and its cached direction is $u^0=U_{p^0}(d)$.
For a coefficient vector $a$ with zeros outside $\mathcal S$, define
\begin{equation}
U_p(a)=a-(\mathbf1^\top a)p,
 \quad N_p(a)=\|a\|_2^2-2(\mathbf1^\top a)p^\top a
 +(\mathbf1^\top a)^2\|p\|_2^2.
\label{eq:app_norm}
\end{equation}
Expansion gives $N_p(a)=\|U_p(a)\|_2^2$.
The support comparison in Section~\ref{sec:app_support_approximation}
also examines a diagnostic scaling against the unsoftened contrast
$d^{\rm raw}_v=\mathbf1[v\in\mathcal S](q^O_v-q^S_v)$.
At its fixed Actor distribution, when both norms are finite and at least
$10^{-24}$, that diagnostic uses
\begin{equation}
 \gamma=\sqrt{\frac{N_p(d^{\rm raw})}{N_p(d^{\rm soft})}},\quad
 d^{\rm diag}=\gamma d^{\rm soft},\quad
 u^{\rm diag}=U_p(d^{\rm diag}).
 \label{eq:app_norm_matching}
\end{equation}
Otherwise the diagnostic retains the finite unsoftened contrast.
This scaling describes the analytic illustration and the recorded
offline analyses. It is absent from the main training coefficient and
from Algorithm~(\ref{alg:app_training}). The diagnostic conventions are
specified alongside the corresponding measurements below.

\begin{figure}[!htb]
\centering
\includegraphics[width=\textwidth]{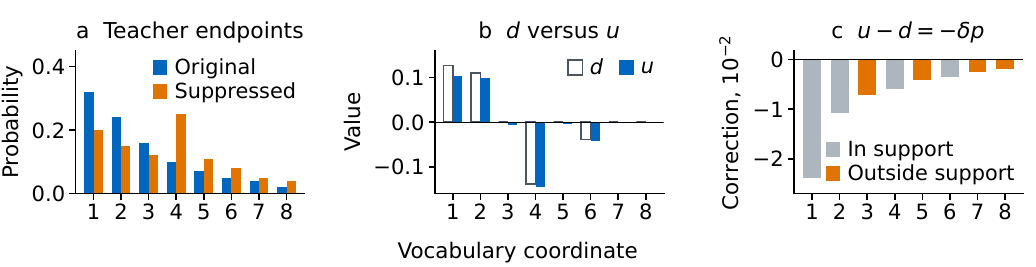}
\caption{\textbf{Analytic illustration with diagnostic scale matching.} With $K=2$ per
endpoint and sampled coordinate 6, the support is $\mathcal S=\{1,2,4,6\}$,
with $M_O=0.71$ and $M_S=0.68$. Panel a shows the teacher endpoints.
Panel b labels the diagnostic coefficient $d^{\rm diag}$ as $d$
in hollow bars and its logit gradient $u^{\rm diag}$ as $u$ in blue. Panel c shows $u-d=-\delta p$ on a
separate scale, with orange outside the support and gray inside it.
With the diagnostic coefficient sum $\delta\approx0.05983$,
the correction extends beyond the support.
The actor distribution is $(.40,.18,.12,.10,.07,.06,.04,.03)$.
The scaling follows Equation~\eqref{eq:app_norm_matching}.
This example isolates how retained mass affects centering.}
\label{fig:app_support}
\end{figure}

\subsection{Exact Gradient of the Signed Objective}
Hold the cached support $\mathcal S$, coefficient $d$, and nonnegative
weight $w$ fixed. For current $p_\theta=\operatorname{softmax}(z)$, define
\begin{equation}
 s(z)=\sum_{v\in\mathcal S}d_v\log p_{\theta,v},\qquad
 \delta=\sum_{v\in\mathcal S}d_v.
 \label{eq:app_signed_score}
\end{equation}
Differentiating $\log p_{\theta,v}$ in $z_j$ gives
$\mathbf1[v=j]-p_{\theta,j}$.
\begin{equation}
\frac{\partial s}{\partial z_j}
 =\sum_{v\in\mathcal S}d_v(\mathbf1[v=j]-p_{\theta,j})
 =d_j-\delta p_{\theta,j},\quad
 \nabla_zs=\widehat u(\theta)=d-\delta p_\theta,\quad
 \mathbf1^\top\widehat u(\theta)=0.
\label{eq:app_signed_gradient}
\end{equation}
At rollout start, $\widehat u(\theta_{\rm old})=u^0$.
Later, $\widehat u(\theta)-u^0=-\delta(p_\theta-p^0)$.
This gives the live loss gradient while preserving the rollout's
cached PVEC coefficients and CRA allocations.
Outside $\mathcal S$, the derivative is $-\delta p_{\theta,j}$.
The loss $-ws$ therefore has gradient $-w\widehat u(\theta)$ before
masking and averaging. It reduces to $-wu^0$ at the initial snapshot.

\subsection{Base Credit and Actual Local Credit}
All credit quantities in this subsection use the rollout snapshot,
with $p=p^0$ and $u=u^0$.
The sampled token score has gradient $v=e_y-p$.
For fixed advantage $A$, the credit used to set the base budget is
\begin{equation}
 b^{\rm base}=A(d_y-p^\top d),\qquad c^0=\operatorname{Var}_p(d).
 \label{eq:app_base_credit}
\end{equation}
The local credit of the visual logit direction is
\begin{equation}
b=A\langle u,e_y-p\rangle
 =A\left[d_y-p^\top d-\delta(p_y-\|p\|_2^2)\right].
 \label{eq:app_local_credit}
\end{equation}
Thus the two credits coincide for zero mass difference, but not in
general. For the separately scaled diagnostic example in
Figure~\ref{fig:app_support}, $A=1$ and $y=6$ give
$\gamma=1.9943$, $\delta=0.059829$, $b^{\rm base}=-0.093162$,
and $b=-0.083122$. These example values use $d^{\rm diag}$ and $u^{\rm diag}$.
Their difference quantifies the centering adjustment to local credit.

The verifier supplies response-level advantages from complete answers.
A positive $b$ identifies local alignment of the visual and
advantage-weighted score gradients. This position-specific support
combines with cross-question credit to guide CRA's allocation.

\subsection{Cross-Question Peer Credit}
For a linear output head $z=Wa$ at fixed hidden input $a$, the chain rule
gives $\nabla_W\log p_y=(e_y-p)a^\top$.
Let $\mathcal O_0$ contain the sampled rollout and let
$N_{\rm tok}^0=\sum_{j\in\mathcal O_0,s}m_{j,s}>0$.
At rollout start the current and sampling policies coincide, so the
likelihood ratio is one. The fold contribution to the globally
normalized GRPO ascent direction is
\begin{equation}
 G_k=\frac{1}{N_{\rm tok}^0}
 \sum_{j:f(j)=k}\sum_s m_{j,s}A_j
 (e_{y_{j,s}}-p^0_{j,s})(a^0_{j,s})^\top.
 \label{eq:fold_gradient}
\end{equation}
The token denominator covers the sampled rollout across workers.
Partitioning it into microbatches does not replace that denominator
with a separate token count for each microbatch.
All valid responses contribute, including those without evidence regions,
and zero advantages contribute zero. Likelihoods use the full softmax.

All responses to one question share one of four deterministic folds.
For an eligible position, define
\begin{equation}
I_{i,t,k}=\frac{\langle G_k,u^0_{i,t}(a^0_{i,t})^\top\rangle_F}
 {\max\{\|G_k\|_F\|a^0_{i,t}\|_2,\epsilon_I\}},
 \quad \mu^{(-f(i))}_{i,t}=\frac13\sum_{k\ne f(i)}I_{i,t,k}.
 \label{eq:app_peer}
\end{equation}
Here $\epsilon_I>0$ is the stabilizer in
Equation~\eqref{eq:peer_influence}.

The denominator omits $\|u^0\|_2$,
so $I$ is not a cosine similarity. A zero head gradient or hidden input
gives zero credit under the same stabilized formula, while the peer
average retains its divisor three. Peer credit is cached with the
remaining correction quantities and measures agreement with the
output head reward directions of other questions.

\subsection{Entropy Derivative and Risk Bounds}
At the rollout snapshot, write $p=p^0$ and $u=u^0$.
For finite logits define $H(p)=-\sum_vp_v\log p_v$. Holding $u$
constant while differentiating entropy,
\begin{equation}
\frac{\partial H}{\partial z_j}
 =-\sum_v p_v(\mathbf1[v=j]-p_j)(\log p_v+1)
 =-p_j\left(\log p_j-\mathbb E_p\log p\right).
\label{eq:app_entropy_gradient}
\end{equation}
The constant terms cancel because $\sum_vp_v=1$. Hence
\begin{equation}
\chi=\nabla_zH^\top u=-\operatorname{Cov}_p(\log p,u),
 \quad F=\operatorname{Var}_p(u),\quad V=\operatorname{Var}_p(\log p).
\label{eq:app_entropy_moments}
\end{equation}
Applying Cauchy--Schwarz to the centered vectors gives $\chi^2\le FV$.
The training risk uses the positive additive stabilizer from
Equation~\eqref{eq:directional_cost},
\begin{equation}
 \rho=\frac{[\min(\chi,0)]^2}
 {(V+\epsilon_\rho)(F+\epsilon_\rho)},\qquad \epsilon_\rho>0.
 \label{eq:app_risk}
\end{equation}
Set $C=\|u\|_2^2(1+\rho)$.
The bound follows directly,
\[
 0\le\rho\le
 \frac{VF}{(V+\epsilon_\rho)(F+\epsilon_\rho)}\le1.
\]
Thus $\|u\|_2^2\le C\le2\|u\|_2^2$.
If either variance is zero, then $\chi=0$ and $\rho=0$ with the
same formula. There is no thresholded branch in the training definition.
The cached risk and cost are reused for every Actor step in that rollout.

The three quantities $c^0=\operatorname{Var}_p(d)$,
$F=\operatorname{Var}_p(u)$, and $C$ have different uses.
The first moderates local credit, the second normalizes directional
entropy contraction, and the third measures allocation cost. Risk
increases the relative cost of a locally entropy-decreasing direction.
This adjustment integrates predictive uncertainty into the allocation
alongside reward credit.

\subsection{Response-Level Allocation}
Define the base and qualified credits by
\begin{equation}
w^0_{i,t}=r_im_{i,t}\frac{[b^{\rm base}_{i,t}]_+}
 {[b^{\rm base}_{i,t}]_++\alpha c^0_{i,t}+\epsilon},\quad
 h_{i,t}=r_im_{i,t}\frac{[b_{i,t}]_+}
 {[b_{i,t}]_++\alpha c^0_{i,t}+\epsilon}
 [\mu^{(-f(i))}_{i,t}]_+,\quad B_i=\sum_t C_{i,t}(w^0_{i,t})^2.
\label{eq:app_credit_budget}
\end{equation}
Let $\mathcal A_i$ contain positions with $r_im_{i,t}=1$,
$h_{i,t}>0$, and $C_{i,t}>0$. Positions outside this set have
zero allocated weight. With $h,C,B_i$ fixed, consider
\begin{equation}
 \max_{w_t\ge0}\sum_{t\in\mathcal A_i}h_tw_t
 \quad\text{subject to}\quad
 \sum_{t\in\mathcal A_i}C_tw_t^2\le B_i.
 \label{eq:app_allocation_problem}
\end{equation}
Weighted Cauchy--Schwarz yields
\begin{equation}
 \sum_t h_tw_t\le
 \sqrt{\sum_t h_t^2/C_t}\sqrt{\sum_t C_tw_t^2}
 \le\sqrt{D_iB_i}.
 \label{eq:app_allocation_bound}
\end{equation}
For $D_i=\sum_{t\in\mathcal A_i}h_t^2/C_t>0$, equality is attained by
\begin{equation}
 w^{\rm alloc}_{i,t}=\sqrt{B_i/D_i}\,h_{i,t}/C_{i,t}
 \quad(t\in\mathcal A_i).
 \label{eq:app_allocation_solution}
\end{equation}
When $B_i=0$, positive costs on $\mathcal A_i$ force zero allocated
weights. If $\mathcal A_i$ is empty, including when all credits vanish,
the imposed zeros outside this set again leave only the zero vector.
This attains the global optimum of the fixed-credit allocation problem.
The resulting weights express correction strength and can exceed one.

\begin{figure}[!htb]
\centering
\begin{minipage}[c]{.49\linewidth}
\centering
\includegraphics[width=\linewidth]{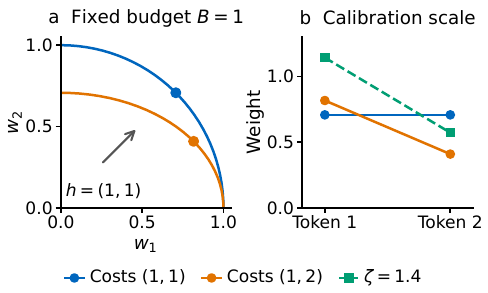}
\end{minipage}\hfill
\begin{minipage}[c]{.48\linewidth}
\captionsetup{justification=raggedright,singlelinecheck=false}
\caption{\textbf{Analytic two-position allocation.} Both credits and the
budget equal one. Blue uses costs $(1,1)$ and orange uses $(1,2)$;
the latter gives $(w_1,w_2)=(0.8165,0.4082)$ and $w_2/w_1=1/2$.
Panel a shows cost boundaries and maximizing allocations. Panel b shows
scaling by $\zeta=1.4$ in green, which preserves relative weights and
multiplies expenditure by $1.4^2$. The fixed budget isolates relative cost.}
\label{fig:app_allocation}
\end{minipage}
\end{figure}

\subsection{Energy Calibration and Averaging}
For position weights $a_{i,t}$ define, at rollout start,
\begin{equation}
\mathcal E(a)=\sum_i\frac{1}{L_i^2}
 \sum_t m_{i,t}a_{i,t}^2\|u^0_{i,t}\|_2^2,
 \quad E_{\rm base}=\mathcal E(w^0),\quad
 E_{\rm alloc}=\mathcal E(w^{\rm alloc}).
\label{eq:calibration_energy}
\end{equation}
A response-normalized visual loss contributes logit gradient
$-m_{i,t}a_{i,t}u^0_{i,t}/L_i$, apart from a common batch denominator
and loss coefficient. Squaring gives the factor $L_i^{-2}$; the common
factors cancel in the energy ratio. The separable sum of squared
position-level logit gradients gives a computable calibration measure;
shared-parameter gradients also involve cross-position Jacobian terms.

Calibration requires finite energies and $E_{\rm alloc}>0$; otherwise
no Actor update is taken. We set
\begin{equation}
 \zeta=\sqrt{E_{\rm base}/E_{\rm alloc}},\qquad
 w^{\rm used}_{i,t}=\operatorname{sg}(\zeta)w^{\rm alloc}_{i,t}.
 \label{eq:calibration_scale}
\end{equation}
Then $\mathcal E(w^{\rm used})=E_{\rm base}$ at calibration. If
$E_{\rm base}=0$, all response budgets and $E_{\rm alloc}$ are zero,
violating the calibration condition.
The calibrated expenditure for response $i$ becomes
$\zeta^2 B_i$ when allocation saturates its budget. Both the multiplier and the allocated weights remain fixed during
subsequent Actor steps. The energy match holds at the rollout snapshot, while
Equation~\eqref{eq:app_signed_gradient} tracks live gradients during later
Actor steps with cached allocations.

\paragraph{Loss Normalization and Gradient Accumulation.}
For an optimizer batch $\mathcal O$, let $N_{\rm aux}=|\mathcal O|$
and $N_{\rm tok}=\sum_{i\in\mathcal O,t}m_{i,t}>0$.
Both counts include all workers.
With current Actor probabilities and cached corrections,
\begin{equation}
s_{i,t}=\sum_{v\in\mathcal S_{i,t}}\operatorname{sg}(d_{i,t,v})
 \log p_{\theta,i,t,v},
 \quad \mathcal L_{\rm vis}=-\frac1{N_{\rm aux}}\sum_{i\in\mathcal O}
 \frac1{L_i}\sum_t m_{i,t}\operatorname{sg}(w^{\rm used}_{i,t})s_{i,t}.
\label{eq:app_visual_loss}
\end{equation}
The response denominator includes ineligible responses, whose visual
summands are zero. Write $\mathfrak B(\mathcal O)$ for any partition into
response microbatches across workers.
The token-averaged policy objective~\citep{ref17} in Section~\ref{sec:grpo} is
\begin{equation}
 \mathcal L_{\rm GRPO}=\frac1{N_{\rm tok}}
 \sum_{\mathcal B\in\mathfrak B(\mathcal O)}
 \sum_{i\in\mathcal B,t}m_{i,t}\ell^{\rm PG}_{i,t}.
 \label{eq:app_pg_average}
\end{equation}
Every microbatch contributes a token sum divided by the same global
denominator, even when microbatch token counts differ.
For $r^{\rm IS}=p_\theta(y)/p_{\theta_{\rm old}}(y)$ and clipping
parameter $\epsilon_{\rm clip}>0$,
\begin{equation}
\bar r^{\rm IS}
 =\operatorname{clip}(r^{\rm IS},1-\epsilon_{\rm clip},
                                  1+\epsilon_{\rm clip}),
 \quad \ell^{\rm PG}
 =-\min\{A r^{\rm IS},A\bar r^{\rm IS}\}.
\label{eq:app_pg_scalar}
\end{equation}
This is the symmetric clipped GRPO loss~\citep{ref4,ref15}.
The EMA reference uses the same global token denominator,
\begin{equation}
 \mathcal L_{\rm ema}=
 \frac1{N_{\rm tok}}\sum_{i\in\mathcal O,t}
 m_{i,t}D_{\rm KL}(q^{\rm EMA}_{i,t}\|p_{\theta,i,t}).
 \label{eq:app_ema_average}
\end{equation}
Before each Actor step the current EMA teacher supplies the original
view reference at temperature one. Its probabilities are then detached.
For fixed normalized $q$, its gradient is
\begin{equation}
 \nabla_zD_{\rm KL}(q\|p)=-q+(\mathbf1^\top q)p=p-q.
 \label{eq:app_kl_gradient}
\end{equation}
The main total objective is
$\mathcal L_{\rm GRPO}+0.5\mathcal L_{\rm vis}+0.1\mathcal L_{\rm ema}$.
Gradient accumulation sums the local numerators using global counts.
If distributed synchronization averages gradients over $R$ workers,
each local contribution is multiplied by $R$ before that averaging.

Visual correction averages response means, whereas GRPO and the EMA
reference average valid tokens.
Two responses of lengths 2 and 8 with token losses 1 and 0 therefore
give response mean $1/2$ and token mean $2/10$.
This difference follows from the objectives and is independent of
microbatch partitioning.

\subsection{Region-Control Diagnostic}
The region-control comparison uses full-vocabulary teacher distributions. At a fixed sampled prefix,
let $q^R$ be the teacher distribution after an equal-area random-region
suppression with negligible overlap with the evidence region. Let
$\mathcal Q$ contain $N_q$ questions and $\mathcal I(x)$ be the nonempty
set of sampled responses to question $x$. Using Jensen-Shannon
divergence~\citep{ref73}, for $e\in\{S,R\}$ define the response means
and question-equal mean by
\begin{equation}
J_i^e=\frac1{L_i}\sum_t m_{i,t}
 D_{\rm JS}(q^O_{i,t},q^e_{i,t}),
 \quad J^e=\frac1{N_q}\sum_{x\in\mathcal Q}
 \frac1{|\mathcal I(x)|}\sum_{i\in\mathcal I(x)}J_i^e.
\label{eq:app_region_js}
\end{equation}
This definition allows unequal numbers of responses per question.
For that diagnostic alone, soften each complete endpoint as
$\widehat q^e=\operatorname{softmax}(\log q^e/2)$ and set
$d^e=\widehat q^O-\widehat q^e$ and $u^e=U_p(d^e)$.
With the categorical Fisher metric~\citep{ref71}
$\mathcal F(p)=\operatorname{diag}(p)-pp^\top$, the angle is
\begin{equation}
 \theta_F=\arccos\frac{(u^S)^\top\mathcal F(p)u^R}
 {\sqrt{(u^S)^\top\mathcal F(p)u^S}
  \sqrt{(u^R)^\top\mathcal F(p)u^R}}.
 \label{eq:app_region_angle}
\end{equation}
It is defined only when both Fisher norms are nonzero. These definitions
describe that offline comparison and do not replace the finite-support
training direction.
\section{Detailed Algorithms and Hyperparameters}
\label{sec:app_algorithms}

Algorithm~(\ref{alg:app_training}) expands the sampling, visual comparison,
allocation, and Actor/teacher updates of Algorithm~(\ref{alg:ravec}).
The detailed steps preserve the same rollout cache and per-batch loss reductions.

\setcounter{algorithm}{0}
\renewcommand{\thealgorithm}{S\arabic{algorithm}}
\begin{algorithm}[!htbp]
\caption{Complete RAVEC Training with a Fixed Rollout Correction Cache}
\label{alg:app_training}
\linespread{1.075}\small
\color{black}
\hypersetup{citecolor=black,urlcolor=black}
\begin{algorithmic}[1]
\REQUIRE Actor $\pi_\theta$, teacher $\pi_\phi$, data and regions, suppressor $\mathcal T$, sampling settings, and $R$ workers.
\REQUIRE $K$, $K_{\rm fold}=4$, $\alpha,\epsilon,\epsilon_\rho,\epsilon_I,\epsilon_{\rm clip},\tau,\lambda_{\rm vis},c_{\rm ema}$, and training budget.
\STATE Initialize the teacher $\phi\leftarrow\theta$ and optimizer.
\REPEAT
\STATE Freeze $\theta_{\rm old}\leftarrow\theta$. Sample grouped responses from the original images.
\STATE Retain full softmax token likelihoods under $\pi_{\theta_{\rm old}}$. Score rewards and compute the group-normalized $A_i$ as in Section~\ref{sec:method}.
\STATE Form $m_{i,t},L_i,r_i$ and assign all responses to a question to the same deterministic fold.
\STATE At rollout start evaluate $p^0,a^0$ and count global $N_{\rm tok}^0$ over all valid response tokens.
\STATE Form the four $G_k$ using Eq.~\eqref{eq:fold_gradient} and the common rollout token denominator.
\STATE Initialize all visual coefficients, directions, credits, costs, and weights to zero at every position.
\FOR{each valid position with $r_i=1$}
\STATE Replay the initial EMA teacher on the original and suppressed views to obtain $q^O,q^S$.
\STATE Construct $\mathcal S=\operatorname{Top}_K(q^O)\cup\operatorname{Top}_K(q^S)\cup\{y\}$ and retained masses $M_O,M_S$.
\STATE Soften at temperature two while preserving retained mass, using Eq.~\eqref{eq:app_softening}.
\STATE Cache $d=\widetilde q^O-\widetilde q^S$, $\delta=\sum_vd_v$, and $u^0=d-\delta p^0$.
\STATE Compute $b^{\rm base}=A_i(d_y-(p^0)^\top d)$, $c^0=\operatorname{Var}_{p^0}(d)$, and $b=A_i\langle u^0,e_y-p^0\rangle$.
\STATE Compute $I_{i,t,k}$ with stabilizer $\epsilon_I$ and $\mu^{(-f(i))}_{i,t}$ by Eq.~\eqref{eq:app_peer}.
\STATE Compute the base weight $w^0$ and qualified credit $h$ by Eq.~\eqref{eq:app_credit_budget}.
\STATE Compute $\chi=-\operatorname{Cov}_{p^0}(\log p^0,u^0)$, $F=\operatorname{Var}_{p^0}(u^0)$, and $V=\operatorname{Var}_{p^0}(\log p^0)$.
\STATE Set $\rho=[\min(\chi,0)]^2/[(V+\epsilon_\rho)(F+\epsilon_\rho)]$ and $C=\|u^0\|_2^2(1+\rho)$.
\ENDFOR
\FOR{each response $i$}
\STATE Set $B_i=\sum_t C_{i,t}(w^0_{i,t})^2$ and $\mathcal A_i=\{t:r_im_{i,t}=1,\ h_{i,t}>0,\ C_{i,t}>0\}$.
\STATE Set $D_i=\sum_{t\in\mathcal A_i}h_{i,t}^2/C_{i,t}$.
\STATE If $D_i>0$, set $w^{\rm alloc}_{i,t}=\sqrt{B_i/D_i}\,h_{i,t}/C_{i,t}$ on $\mathcal A_i$. Retain zero weights elsewhere.
\ENDFOR
\STATE Compute $E_{\rm base}$ and $E_{\rm alloc}$ over the entire rollout using Eq.~\eqref{eq:calibration_energy}.
\STATE Abort before updating the Actor if energies are nonfinite or $E_{\rm alloc}=0$.
\STATE Set $\zeta=\sqrt{E_{\rm base}/E_{\rm alloc}}$ and $w^{\rm used}=\operatorname{sg}(\zeta)w^{\rm alloc}$.
\STATE Detach and cache $d,u^0,w^{\rm used}$ and all PVEC and CRA quantities for this rollout.
\FOR{each Actor optimization step with response batch $\mathcal O$}
\STATE Count global $N_{\rm aux}=|\mathcal O|$ and $N_{\rm tok}=\sum_{i\in\mathcal O,t}m_{i,t}>0$ across workers.
\STATE Reset accumulated gradients. Recompute and detach the original view EMA reference at temperature one.
\FOR{each local response microbatch $\mathcal B\subset\mathcal O$}
\STATE Evaluate current $p_\theta$ and symmetric clipped token losses $\ell^{\rm PG}_{i,t}$ from Eq.~\eqref{eq:app_pg_scalar}.
\STATE Evaluate $s_{i,t}=\sum_{v\in\mathcal S_{i,t}}d_{i,t,v}\log p_{\theta,i,t,v}$ using the cached coefficient.
\STATE $\ell^{\mathcal B}_{\rm GRPO}=N_{\rm tok}^{-1}\sum_{i\in\mathcal B,t}m_{i,t}\ell^{\rm PG}_{i,t}$.
\STATE $\ell^{\mathcal B}_{\rm vis}=-N_{\rm aux}^{-1}\sum_{i\in\mathcal B}L_i^{-1}\sum_t m_{i,t}w^{\rm used}_{i,t}s_{i,t}$.
\STATE $\ell^{\mathcal B}_{\rm ema}=N_{\rm tok}^{-1}\sum_{i\in\mathcal B,t}m_{i,t}D_{\rm KL}(q^{\rm EMA}_{i,t}\|p_{\theta,i,t})$.
\STATE Accumulate $R\nabla_\theta(\ell^{\mathcal B}_{\rm GRPO}+\lambda_{\rm vis}\ell^{\mathcal B}_{\rm vis}+c_{\rm ema}\ell^{\mathcal B}_{\rm ema})$.
\ENDFOR
\STATE Average gradients across workers, clip the gradient norm, and take the Actor step.
\STATE Update $\phi\leftarrow\tau\phi+(1-\tau)\theta$. Preserve the rollout correction cache.
\ENDFOR
\UNTIL{the training budget is exhausted}
\RETURN Trained Actor $\pi_\theta$.
\end{algorithmic}
\end{algorithm}

\paragraph{Update Order.}
Each rollout follows five stages.
\begin{enumerate}[leftmargin=*,nosep,topsep=3pt,midpenalty=10000]
\item \textbf{Sampling and Scoring.} Sample grouped responses, compute
rewards and advantages, and retain token likelihoods under the sampling policy.
\item \textbf{Visual Comparison.} Replay the initial teacher on paired
views and construct the support, visual coefficient, and snapshot direction.
\item \textbf{Credit and Calibration.} Compute local and peer credits,
stabilized entropy risk, allocation, and the energy multiplier once.
Cache the resulting visual coefficients and final correction weights.
\item \textbf{Actor Update.} Recompute current loss probabilities and
the detached original view EMA reference. Use the cached visual
coefficient and weight without refreshing credit, risk, or allocation.
\item \textbf{Teacher Update.} Update the teacher after the Actor step.
Repeat the final two stages for the remaining optimization steps,
then replace the correction cache when a new rollout is sampled.
\end{enumerate}

The gradient of the visual objective is evaluated through current
probabilities as specified in Equation~\eqref{eq:app_signed_gradient}.
Cached coefficients support optimization with live Actor logits.

\subsection{Training and Evaluation Settings}
\label{sec:app_training_settings}
Table~\ref{tab:app_common_settings} summarizes the main training protocol
and independent studies used for training curves and task diagnostics.
All studies use AdamW~\citep{ref74}.
The smaller studies characterize the fixed main design.
The main configuration follows the fixed cache and stabilized risk in
Algorithm~(\ref{alg:app_training}).
The independent studies retain their stated sampling settings and
reported diagnostic conventions, which are not additional steps in
that training algorithm.

The paired-response analysis uses the training-curve model. Task-wise
and fixed-prefix analyses use the independent diagnostic model.
Model parameters remain fixed during these offline analyses.

\paragraph{Model and Data Sources.}
The Qwen2.5-VL~\citep{qwen25vl2025} models train on 38,870 ViRL39K
questions from VL-Rethinker~\citep{virl39k2025}, following PAPO~\citep{ref56}.
Benchmark sources are Inter-GPS for Geometry3K~\citep{geometry3k2021},
MathVista~\citep{mathvista2024}, We-Math~\citep{wemath2025},
MathVerse~\citep{mathverse2024}, LogicVista~\citep{logicvista2024},
MMMU-Pro~\citep{mmmupro2025}, C-VQA-Real for CVQA~\citep{cvqa2024},
MARS-Bench for MARS~\citep{mars2025}, and POPE~\citep{pope2023}.
Counting uses Super-CLEVR~\citep{superclevr2023} and \mbox{MathVerse-V} uses
MathVerse; both follow the PAPO~\citep{ref56} subset definitions.

\begin{table}[!htb]
\centering\normalsize
\renewcommand{\arraystretch}{1.18}
\caption{\textbf{Training settings.} Studies use Qwen2.5-VL-3B-Instruct,
seed 1, and one pass over sampled responses per rollout. Update batches
count responses. Training and validation sampling are listed separately.}
\label{tab:app_common_settings}
\setlength{\tabcolsep}{3pt}
\begin{tabular}{@{}L{.32\linewidth}L{.22\linewidth}L{.20\linewidth}L{.20\linewidth}@{}}
\toprule
Setting & Main 3B GRPO & Six-method curves & Task diagnostics \\
\midrule
Training questions & 38,870 & 1,024 & 1,024 \\
Validation questions & 2,000 & 160 & 160 \\
Sampled batches & 202 & 200 & 200 \\
Optimizer updates & 606 & 200 & 200 \\
Questions per sampled batch & 384 & 32 & 16 \\
Training answers per question & 5 & 8 & 8 \\
Responses per sampled batch & 1,920 & 256 & 128 \\
Responses per optimizer update & 640 & 256 & 128 \\
Response microbatch per device & 2 & 2 & 2 \\
Validation times & 10--100 every 10; 150,175 & 0--200 every 20 & 0--200 every 20 \\
Optimizer & AdamW & AdamW & AdamW \\
Learning rate / schedule & $10^{-6}$ / constant & $10^{-6}$ / constant & $10^{-6}$ / constant \\
Warmup; $(\beta_1,\beta_2)$ & 0; $(.9,.999)$ & 0; $(.9,.999)$ & 0; $(.9,.999)$ \\
Weight decay; gradient clipping & .01; 1.0 & .01; 1.0 & .01; 1.0 \\
Maximum input / output tokens & 16,384 / 2,048 & 16,384 / 2,048 & 16,384 / 2,048 \\
Training temperature / top-$p$ & 1 / 1 & 1 / .99 & 1 / .99 \\
Validation temperature / top-$p$ & 1 / 1 & 1 / 1 & 1 / 1 \\
Validation generations per question & 8 & 8 & 8 \\
Sampling top-$k$ cutoff & None & None & None \\
Image pixel bounds & 200,704--1,003,520 & 200,704--1,003,520 & 200,704--1,003,520 \\
Parameter / forward precision & FP32 / BF16 & FP32 / BF16 & FP32 / BF16 \\
\bottomrule
\end{tabular}
\end{table}

\subsection{RAVEC Parameters}

The budget $B_i$ and calibration multiplier $\zeta$ are computed
and cached once per rollout.
The positive constants $\epsilon_\rho$ and $\epsilon_I$ and the symmetric
clipping parameter $\epsilon_{\rm clip}$ retain their definitions in
Equations~\eqref{eq:directional_cost}, \eqref{eq:peer_influence}, and
\eqref{eq:app_pg_scalar}.

\begin{table}[!htb]
\centering\normalsize
\renewcommand{\arraystretch}{1.18}
\caption{\textbf{RAVEC parameters.} The independent training-curve and
diagnostic studies share
the listed changes from the main configuration.}
\label{tab:app_ravec_settings}
\begin{tabular}{@{}L{.13\linewidth}L{.31\linewidth}L{.13\linewidth}L{.34\linewidth}@{}}
\toprule
Symbol & Meaning & Main 3B & Training-curve and diagnostic studies \\
\midrule
$\lambda_{\rm vis}$ & Visual loss coefficient & 0.5 & 0.75 \\
$K$ & Candidates per teacher endpoint & 128 & Unchanged \\
$\tau_v$ & Mass-preserving softening temperature & 2 & Unchanged \\
$\alpha$ & Base-direction cost coefficient & 0.0025 & Unchanged \\
$\epsilon$ & Continuous-credit \mbox{denominator constant} & $2.5\times10^{-5}$ & $10^{-8}$ \\
$\tau$ & Teacher exponential decay & 0.95 & Unchanged \\
$K_{\rm fold}$ & Question folds for peer credit & 4 & Unchanged; all siblings remain in one fold \\
--- & Evidence-region linear resize ratio & 0.1 & Unchanged \\
$c_{\rm ema}$ & Full-vocabulary reference coefficient & 0.1 & 0.1 in the GRPO studies \\
$T_{\rm ema}$ & Reference-distribution temperature & 1 & Unchanged \\
\bottomrule
\end{tabular}
\end{table}

\setcounter{topnumber}{1}
\setcounter{bottomnumber}{1}
\renewcommand{\floatpagefraction}{.85}
\Needspace{8\baselineskip}
\section{Lightweight Diagnostics}
\label{app:diagnostics}

\subsection{Finite-Support Approximation}
\label{sec:app_support_approximation}

With fixed Actor and teacher, we compare finite-support and full-vocabulary
softening at answer prefixes in Section~\ref{sec:app_probes}, regardless
of correctness. These concentrated answer distributions provide a focused
test of final-decision approximation. Let $T$ denote the softening temperature $\tau_v$ in
Eq.~\eqref{eq:app_softening}.

We use $T=2$ as the primary comparison and
$T\in\{1,4\}$ as additional settings. For $K\in\{64,128,256\}$, the
support is the union in Eq.~\eqref{eq:app_support}, including the actual
sampled answer token. Write $\bar u_{K,T}$ for the mass-preserving softened gradient
before norm matching. The full-vocabulary reference uses
$q^e_T=\operatorname{softmax}(z^e/T)$ for both endpoints and
$\bar u_{\infty,T}=U_p(q^O_T-q^S_T)$, with the same fixed Actor
distribution $p$. Both gradients include every coordinate of
$U_p(a)=a-(\mathbf1^\top a)p$; the finite-support comparison therefore
retains the support-exterior normalization term.
Full-vocabulary softening is a diagnostic reference here; it is not the
finite-support direction used in training.

We apply Eq.~\eqref{eq:app_norm_matching} separately and identically
to each support: $u_{K,T}$ matches the norm of its own unsoftened
gradient $\bar u_{K,1}$, and $u_{\infty,T}$ matches
$\bar u_{\infty,1}$. Both comparisons use the same diagnostic norm matching rule.
Only the unscaled coefficient $d=d^{\rm soft}$ enters the main
training algorithm. The before and after comparisons here isolate
the effect of the optional diagnostic scaling.
We report cosine similarity and relative errors to assess
direction and amplitude
\begin{equation}
R_{\rm pre}
 =\frac{\|\bar u_{K,T}-\bar u_{\infty,T}\|_2}
 {\|\bar u_{\infty,T}\|_2},
 \quad R_{\rm post}
 =\frac{\|u_{K,T}-u_{\infty,T}\|_2}
 {\|u_{\infty,T}\|_2}.
\label{eq:app_support_errors}
\end{equation}
Both metrics are defined for the directions compared here.

\begin{figure}[tb]
\centering
\includegraphics[width=\textwidth]{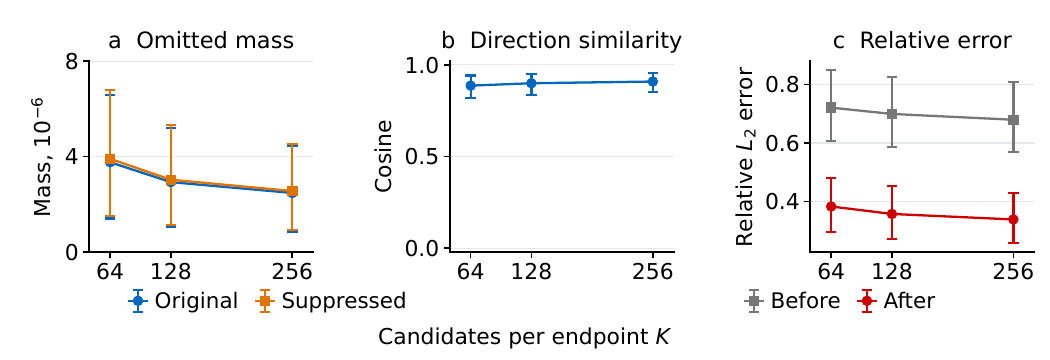}
\caption{\textbf{Finite-support approximation at $T=2$.}
Panel a shows omitted original-temperature mass $10^6(1-M_e)$;
blue and orange denote original and suppressed teacher views.
Panel b compares norm-matched logit directions by cosine similarity.
Panel c shows relative $L_2$ errors against full-vocabulary softening,
in gray before norm matching and red afterward. Points and bars give
question means and 95\% task-stratified paired bootstrap intervals.
Lines join the measured support sizes.}
\label{fig:app_support_approximation}
\end{figure}

At $T=2,K=128$, the support retains over 99.999\% of the
original-temperature mass at both endpoints, with mean cosine 0.8999
and post-matching relative error 0.3571. The 95\% question-bootstrap
interval~\citep{bootstrap1979} is $[0.2725,0.4521]$. Larger $K$ further reduces mean error.
Figure~\ref{fig:app_support_approximation} jointly characterizes mass retention and directional
agreement with the full-vocabulary diagnostic reference.

At $T=1$, mean relative error stays below 0.006. At $T=4$,
norm matching reduces amplitude error; position-level $R_{\rm post}$
values quantify directional and amplitude variation.

\subsection{What the Lightweight Analyses Measure}
Fixed-prefix probes characterize conditional answer distributions. The
support comparison quantifies approximation at selected answer positions.
Together, they complement end-to-end accuracy with focused mechanism tests.

Using the full-vocabulary direction as a diagnostic reference isolates
the effects of support restriction and diagnostic scaling. Approximation
error measures fidelity to that reference, while the main benchmark
evaluation assesses the resulting framework's answer accuracy.

\section{Training and Token-Level Analyses}
\label{app:training}

\subsection{Six-Method Training Dynamics}
The six methods share initialization, training settings, and validation
questions, as specified in Table~\ref{tab:app_common_settings}.
Tasks contribute equally to validation, so task-equal and
question-weighted averages coincide. This controlled comparison uses
a common protocol rather than reproducing each method's original
training recipe.

\begin{figure}[tb]
\centering
\includegraphics[width=\textwidth]{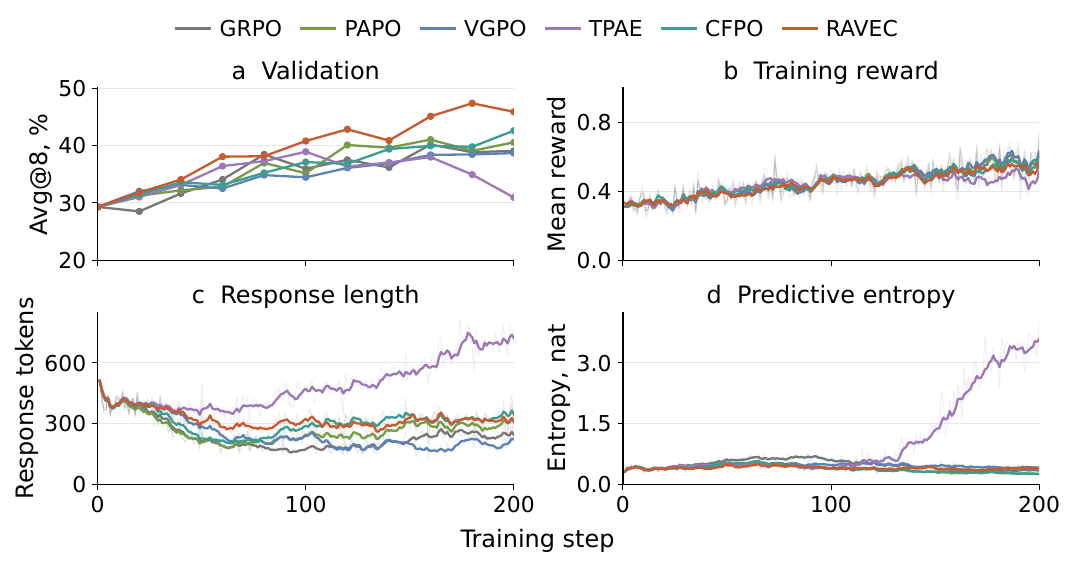}
\caption{\textbf{Six-method training dynamics.} Colors identify methods.
Validation reports Avg@8; reward averages answer and format scores,
length counts response tokens, and entropy averages full-vocabulary
predictive entropy over valid tokens. Pale curves show raw training
values and dark curves show trailing five-step means.}
\label{fig:app_training}
\end{figure}

Figure~\ref{fig:app_training} separates validation accuracy from training
reward, response length, and predictive entropy. Training reward is
$0.9$ times answer correctness plus $0.1$ times format validity.
Predictive entropy is the full-vocabulary entropy before the update,
averaged over valid generated response tokens. These two quantities
therefore have different statistical units and meanings from validation
Avg@8. Smoothing is used only for display.

Table~\ref{tab:app_training_endpoints} distinguishes the common final
evaluation from the highest observed score of each trajectory. The final RAVEC--CFPO
difference is 3.28 percentage points, with a 95\%
task-stratified question-bootstrap interval~\citep{bootstrap1979} of $[-0.08,6.80]$ points.
Resampling keeps each question's responses together and conditions
on the trained models.

The interval summarizes differences across resampled question sets.

\begin{figure}[tb]
\centering
\includegraphics[width=\textwidth]{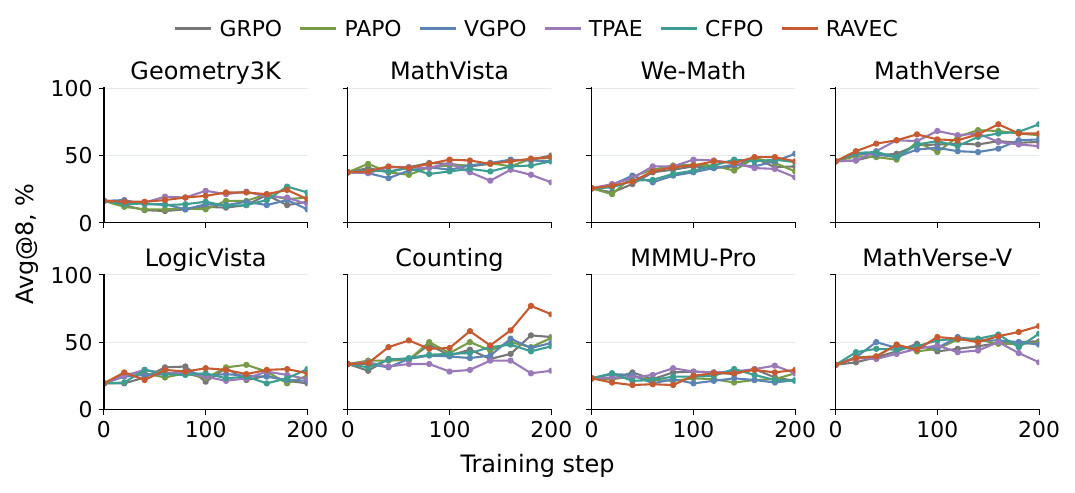}
\caption{\textbf{Validation by task.} The same methods and validation
set as Figure~\ref{fig:app_training}. Vertical axes show Avg@8 in
percent on a common scale. Colors identify methods, and markers show
measured evaluations.}
\label{fig:app_task_validation}
\end{figure}

Figure~\ref{fig:app_task_validation} shows all eight task trajectories,
complementing aggregate accuracy with task-specific progress. Resampling
compares the fixed models, with questions as the units of
sampling~\citep{rl_statistics2021}.

{\clubpenalty=10000
RAVEC reaches its highest observed score before the final checkpoint,
whereas CFPO and VGPO peak at the final evaluation. Peak and final
scores therefore describe different aspects of the trajectories.

The common final checkpoint avoids selecting a separate
evaluation time for each method.\par}

\begin{table}[tb]
\centering
\small
\setlength{\tabcolsep}{4pt}
\begin{minipage}[c]{.66\linewidth}
\centering
\begin{tabular}{lrrr}
\toprule
Method & Final & Highest observed & Peak step \\
\midrule
GRPO~\citep{ref15} & 39.06 & 40.16 & 160 \\
PAPO~\citep{ref56} & 40.55 & 41.02 & 160 \\
VGPO~\citep{ref79} & 38.67 & 38.67 & 200 \\
TPAE~\citep{ref81} & 30.94 & 38.91 & 100 \\
CFPO~\citep{ref80} & 42.58 & 42.58 & 200 \\
RAVEC & 45.86 & 47.34 & 180 \\
\bottomrule
\end{tabular}
\end{minipage}\hfill
\begin{minipage}[c]{.31\linewidth}
\captionsetup{justification=raggedright,singlelinecheck=false}
\caption{\textbf{Final and highest observed validation scores.}
Values are Avg@8 in percent. Final scores use a common checkpoint;
peak scores summarize each training trajectory.}
\label{tab:app_training_endpoints}
\end{minipage}
\end{table}

\subsection{Task-Wise Correction Statistics}
\label{sec:app_recorded_diagnostics}
The correction diagnostics use the independent training run listed in
Table~\ref{tab:app_common_settings}. Held-out questions remain fixed,
but responses are sampled anew at each evaluation. Trends therefore
reflect both changes in the model and changes in generated content.

We define $\mathcal E_{s,d,c}$ as the eligible token events at step $s$
from task $d$ and correctness stratum $c$. Eligibility requires a valid
response position, an available evidence region, and a nonzero response
advantage; the plotted quantities must be finite.

The strata are all
responses, correct responses, and incorrect responses.

Eligible
zero-weight tokens remain in the denominator; empty sets have
undefined means.

For any token quantity $x$, define its pooled mean by
\begin{equation}
 \langle x\rangle_{s,d,c}
 =\frac{\sum_{(i,t)\in\mathcal E_{s,d,c}}x_{i,t}}
 {|\mathcal E_{s,d,c}|},
 \qquad |\mathcal E_{s,d,c}|>0.
 \label{eq:app_token_pool}
\end{equation}
Table~\ref{tab:app_diagnostic_definitions} specifies the four displayed
quantities. Pooling sums the numerators and denominators across
responses; it is not an unweighted average of response-level means.

Panel b of Figure~\ref{fig:app_peer} tracks eligibility across training.
Token counts also depend on response length.

Responses contribute in proportion to their eligible token counts.
Changes in pooled credit or risk can therefore reflect both the
per token quantities and the composition of sampled responses.
The curves should be read together with the eligibility counts in
Figure~\ref{fig:app_peer}.

\begin{figure}[tb]
\centering
\includegraphics[width=\textwidth]{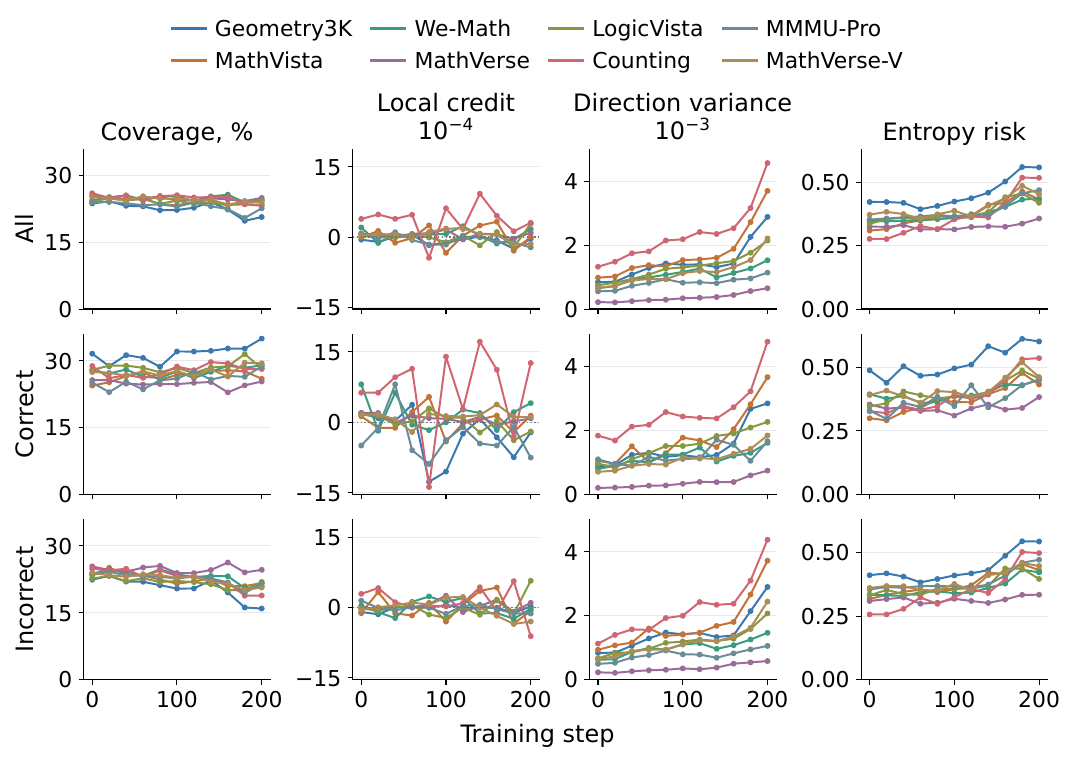}
\caption{\textbf{Correction dynamics by task and response correctness.}
Colors identify tasks; rows show all, correct, and incorrect responses.
Means pool eligible tokens, including zero weights, using the
denominators in Table~\ref{tab:app_diagnostic_definitions}.
Credit and direction variance are scaled by $10^4$ and $10^3$ for display.
Axes are shared across correctness strata within each column.}
\label{fig:app_token_dynamics}
\end{figure}

The recorded offline analyses use the scale matching in
Equation~\eqref{eq:app_norm_matching}.

Within these analyses, $d$ and $u$
denote $d^{\rm diag}$ and $u^{\rm diag}$, and credits and weights are
evaluated using these diagnostic directions with the model held fixed.
Their recorded entropy statistic uses the zero stabilizer limit
where the variances exceed numerical cutoffs,
\[
 \rho_{\rm diag}=\begin{cases}
 [\min(\chi,0)]^2/(VF),&V>10^{-24},\ F>10^{-30},\\
 0,&\text{otherwise}.
 \end{cases}
\]
These conventions preserve the reported diagnostic measurements.
They do not replace the unscaled coefficient or the positive additive
stabilizer of the main training algorithm. In particular,
$\rho_{\rm diag}$ need not equal the stabilized training risk in
Equation~\eqref{eq:app_risk}.

Figure~\ref{fig:app_token_dynamics} reports each task and correctness
stratum. Signed credit is measured before positive clipping, so negative
means remain visible. Direction variance is computed before token
weighting and global energy calibration. Entropy risk is dimensionless;
it differs from the entropy itself and from the complete allocation
cost. The plots characterize correction dynamics across tasks and
training, complementing benchmark accuracy.

\begin{table}[tb]
\centering
\small
\begin{tabular}{L{.20\linewidth}L{.38\linewidth}L{.34\linewidth}}
\toprule
Quantity & Definition & Unit and denominator \\
\midrule
Effective coverage & $100\langle\mathbf 1[w^{\mathrm{used}}>10^{-3}]\rangle_{s,d,c}$ & Percent of eligible finite token events; eligible zeros are included. \\
Signed local credit & $\langle b\rangle_{s,d,c}$, with $b=A\,u^\top(e_y-p)$ & Credit per eligible token; no positive clipping. \\
Direction variance & $\langle F\rangle_{s,d,c}$, with $F=\operatorname{Var}_p(u)$ & Variance per eligible token, before weighting. \\
Diagnostic entropy risk & $\langle\rho_{\rm diag}\rangle_{s,d,c}$ & Dimensionless risk fraction per eligible token, including zeros. \\
\bottomrule
\end{tabular}
\caption{\textbf{Definitions for the checkpoint diagnostics.}
The expectation notation denotes the pooled average in
Equation~\eqref{eq:app_token_pool}, not an average across independent
training runs. Empty eligible sets have an undefined mean.}
\label{tab:app_diagnostic_definitions}
\end{table}

\subsection{Cross-Question Peer Selection}
\label{app:peer_selection}
\begin{figure}[tb]
\centering
\includegraphics[width=\textwidth]{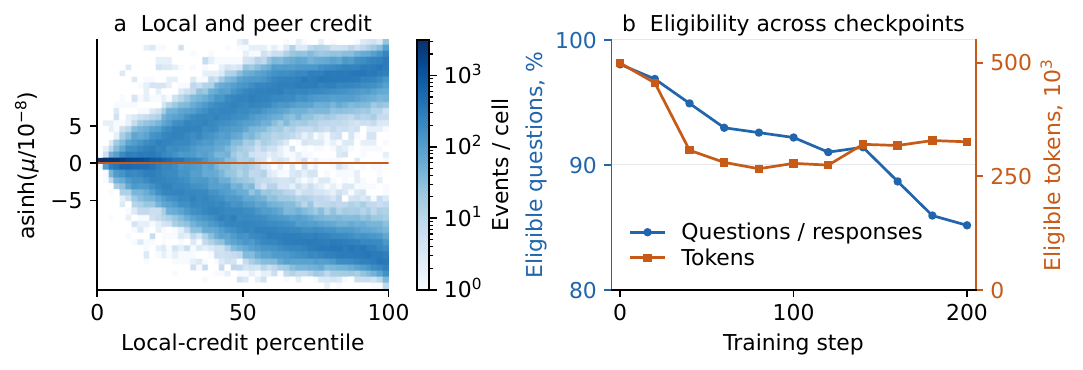}
\caption{\textbf{Eligibility and peer credit.}
Panel a shows positive-local-credit positions by the empirical percentile
of $b$ and $\operatorname{asinh}(\mu/10^{-8})$. Blue intensity gives
counts per cell on a logarithmic scale; the orange line marks $\mu=0$.
In panel b, blue circles show the percentage of questions with at least
one eligible position on the left axis. Orange squares show eligible
token counts in thousands on the right axis. Lines join measured
evaluations.}
\label{fig:app_peer}
\end{figure}

We examine peer selection in the correction-diagnostic run. Candidate
positions have an available evidence region and nonzero response
advantage. Panel a of Figure~\ref{fig:app_peer} shows positions with
positive local credit at the final evaluation. Peer credit is nonpositive
for about half of these positions:
positive local credit therefore does not imply positive peer credit.
The peer test adds gradient-based selection, qualifying locally aligned
corrections by cross-question support.

\subsection{Paired Allocation by Response Correctness}
\label{app:paired_allocation}

The paired analysis uses the model from the controlled training
comparison, independently of the correction-diagnostic run. Each pair
contains a correct and an incorrect response to the same question.
Responses have valid answer formatting and are not truncated; selection
does not use their token weights. Allocation is evaluated with the model
held fixed, rather than measured during generation. The recorded weights
follow the diagnostic conventions in
Section~\ref{sec:app_recorded_diagnostics}.

Let $w_{it}^{\mathrm{used}}$ be the calibrated weight and $L_i$ the number
of valid response tokens, including formatting positions.

We define the
response mean, sum, and effective coverage as
\begin{gather}
 \overline w_i=\frac{1}{L_i}\sum_t m_{it}w_{it}^{\mathrm{used}},\quad
 S_i=\sum_t m_{it}w_{it}^{\mathrm{used}},\label{eq:app_pair_weights}\\
 c_i=\frac{\sum_t m_{it}\mathbf1[w_{it}^{\mathrm{used}}>10^{-3}]}{L_i}.
 \label{eq:app_pair_coverage}
\end{gather}

{\interlinepenalty=10000
The descriptive threshold in $c_i$ is distinct from the positivity tests
used for credit qualification. Means include all valid zero weights,
whether due to ineligibility or zero allocation.\par}

Within each outcome group, the pooled mean is
$\sum_i S_i/\sum_i L_i$, whereas response sums are averaged equally
across responses. Table~\ref{tab:app_pair_summary} uses paired
question-bootstrap intervals~\citep{bootstrap1979} without task
stratification, recomputing
the pooled denominators in each resample. 

\begin{table}[tb]
\centering\small
\caption{\textbf{Paired allocation to correct and incorrect responses.}
Weight and coverage pool valid tokens; response sums and lengths
average responses. Intervals are 95\% paired question-bootstrap
intervals. Coverage and zero-weight differences are in percentage points.}
\label{tab:app_pair_summary}
\begin{tabular}{lrrrr}
\toprule
Metric & Correct & Incorrect & Difference & 95\% interval for difference \\
\midrule
Mean token weight & 224.38 & 93.49 & 130.89 & $[-46.53,\ 476.11]$ \\
Weight sum per response & 69,770.53 & 28,117.53 & 41,653.00 & $[-13,401.46,\ 142,617.63]$ \\
Effective coverage, \% & 21.70 & 22.66 & $-0.96$ & $[-3.02,\ 1.14]$ \\
Exactly zero weight, \% & 77.41 & 76.97 & 0.44 & $[-1.86,\ 2.66]$ \\
Response length, tokens & 310.95 & 300.77 & 10.18 & $[-26.31,\ 47.57]$ \\
\bottomrule
\end{tabular}
\end{table}

The pooled means show greater allocation to correct responses in this
sample, with 64.38\% of their total weight concentrated on one token.

The token-level view resolves how correction strength is distributed.
Question-wise ordering varies; Table~\ref{tab:app_pair_summary} reports paired
uncertainty. Together, these summaries characterize sampled allocation
and its variation across questions.

\begin{figure}[tb]
\centering
\includegraphics[width=\textwidth]{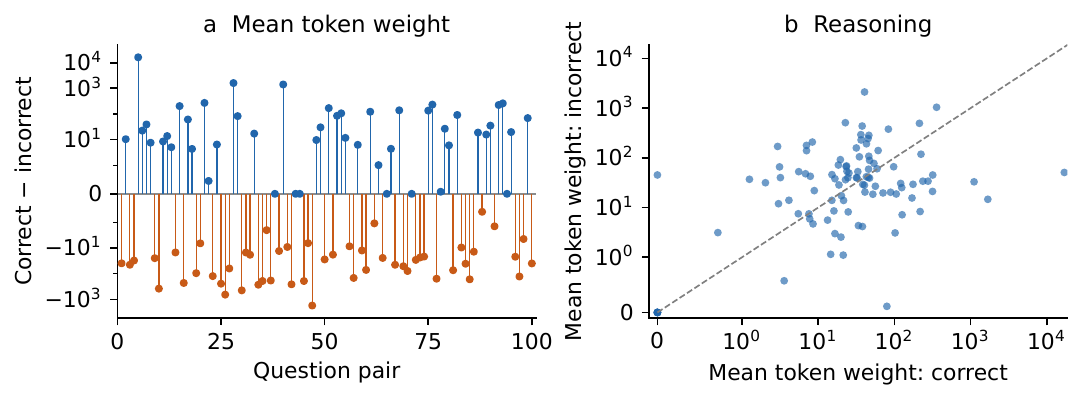}
\caption{\textbf{Paired allocation by response correctness.}
Panel a shows correct-minus-incorrect differences in response-mean
weight. Blue and orange indicate nonnegative and negative differences;
the symmetric-log scale is linear within $\pm1$.
Panel b compares reasoning-token means, with correct responses on the
horizontal axis and incorrect responses on the vertical axis.
The dashed line marks equality; both symmetric-log axes are linear
from zero to one. Means include zero-weight tokens. Response-level
differences need not average to the pooled token-weighted difference
in Table~\ref{tab:app_pair_summary}.}
\label{fig:app_pair_allocation}
\end{figure}

\paragraph{Response Regions.}
Panel b of Figure~\ref{fig:app_pair_allocation} isolates the text inside
the response's reasoning markers. The final-answer continuation and
formatting boundaries are excluded from this panel but remain in the
whole-response means.

\paragraph{Frequently Observed Token Types.}
Table~\ref{tab:app_common_tokens} compares high-weight alphabetic token
types that occur frequently in both response groups, excluding formatting
and mathematical control words. Function words also receive allocation
within generated reasoning. Occurrence means describe lexical patterns;
question-level pairs provide the sampling unit for uncertainty.

\begin{table}[tb]
\centering\small
\caption{\textbf{The same token types in both response groups.}
Counts refer to qualifying alphabetic occurrences, and means include
zero weights. Capitalization distinguishes ``The'' from ``the''.}
\label{tab:app_common_tokens}
\normalsize
\setlength{\tabcolsep}{4pt}
\renewcommand{\arraystretch}{1.08}
\begin{tabular*}{\columnwidth}{@{\extracolsep{\fill}}lrrrr@{}}
\toprule
 & \multicolumn{2}{c}{Correct} & \multicolumn{2}{c}{Incorrect} \\
Token & Count & Mean & Count & Mean \\
\midrule
by & 67 & 3,321.44 & 75 & 2,103.72 \\
the & 977 & 450.95 & 1,047 & 72.17 \\
in & 125 & 397.80 & 122 & 167.70 \\
The & 81 & 233.63 & 84 & 2.18 \\
and & 351 & 198.05 & 349 & 247.25 \\
this & 26 & 1.75 & 37 & 857.84 \\
that & 100 & 60.95 & 133 & 258.69 \\
of & 511 & 123.60 & 510 & 224.75 \\
\bottomrule
\end{tabular*}
\end{table}
\begingroup
\setlength{\abovedisplayskip}{6pt plus 2pt minus 2pt}
\setlength{\belowdisplayskip}{6pt plus 2pt minus 2pt}
\setlength{\abovedisplayshortskip}{0pt plus 2pt}
\setlength{\belowdisplayshortskip}{3pt plus 1pt minus 1pt}
\Needspace{6\baselineskip}
\section{Offline Mechanism Probes}
\label{app:probes}
\label{sec:app_probes}

\setcounter{dbltopnumber}{1}
\begin{figure}[tb]
\centering
\includegraphics[width=\textwidth]{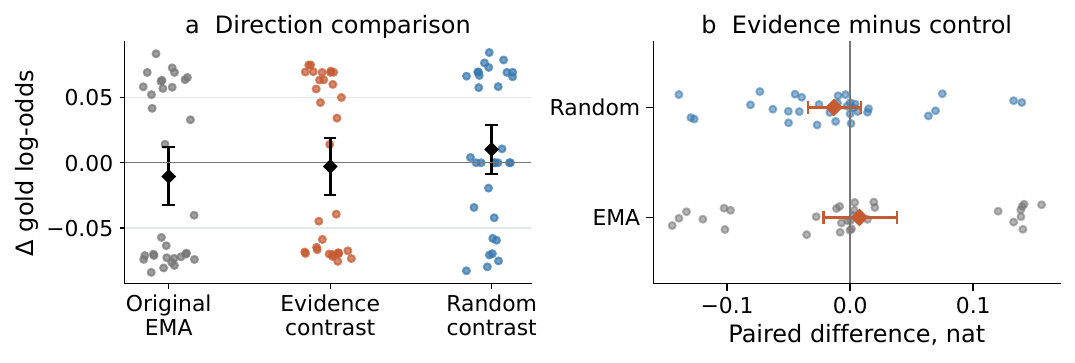}
\caption{\textbf{Fixed-prefix direction probe.} Panel a shows
changes in gold-option log-odds under directions matched to centered
answer-logit RMS 0.03 nat. Panel b shows paired evidence-minus-control
differences. Points represent questions, with jitter for visibility.
Diamonds and bars give means and 95\% task-stratified paired
question-bootstrap intervals. Colors identify directions or controls;
directions below the RMS threshold produce zero perturbations.}
\label{fig:app_direction_probe}
\end{figure}

\subsection{Direction Quality at a Fixed Answer Prefix}
We compare answer-logit directions using the correction-diagnostic
model and existing prefixes immediately before the final answer letter.
Questions have four letter-labelled options and are selected without
conditioning on correctness. Fixed prefixes preserve the generated
reasoning and answer context for a controlled comparison.

Let $\mathcal C=\{A,B,C,D\}$ and let $z_{\mathcal C}$ be the Actor's
four answer logits. The original-image EMA direction is
$(q^O-p)_{\mathcal C}$. The evidence and random-region directions are
the corresponding finite-support diagnostic directions $u_{\mathcal C}$,
with scale matching as specified in
Section~\ref{sec:app_recorded_diagnostics}.
The random region lies in the same image, has the same width and height
as the annotated evidence rectangle, and has zero overlap with it;
source selection requires such a region to exist.
For each $v\in\mathbb R^4$, subtract its coordinate mean and define
\begin{equation}
 \bar v=v-\tfrac14(\mathbf1^\top v)\mathbf1,
 \qquad s(v)=\sqrt{\tfrac14\|\bar v\|_2^2}.
 \label{eq:app_choice_rms}
\end{equation}
The virtual answer logits are $z_{\mathcal C}+\eta\bar v/s(v)$ when
$s(v)>10^{-10}$, and remain unchanged otherwise. Thus the nonzero
directions have matched centered answer-logit RMS. The prespecified
primary strength is $\eta=0.03$ nat; additional strengths are 0.01 and
0.1 nat. The endpoint is the change in gold-option log-odds,
\begin{equation}
 \Delta\ell
 =\left[z'_g-\log\sum_{c\ne g}e^{z'_c}\right]
 -\left[z_g-\log\sum_{c\ne g}e^{z_c}\right].
 \label{eq:app_gold_logodds}
\end{equation}
These are option-conditional quantities, not probabilities normalized
over the entire vocabulary.
Figure~\ref{fig:app_direction_probe} compares the directions at the
primary strength. Probe intervals use task-stratified paired
question bootstrapping~\citep{bootstrap1979}.

\subsection{Virtual Output-Head Transfer}
The second probe constructs virtual output-head changes from the same
source responses. Separate unlabeled questions calibrate perturbation
strength, and held-out questions evaluate transfer. Calibration and
evaluation data are disjoint from each other and from the source and
training data, including their images. These virtual changes alter only
the output logits; the backbone and model weights remain fixed.

We compare four allocation rules: all positive-local-credit positions;
the same number of highest-local-credit positions per response as pass
the peer test; positive peer credit with the continuous local gate;
and the latter rule with the recorded diagnostic entropy risk factor
$\rho_{\rm diag}$ defined in Section~\ref{sec:app_recorded_diagnostics}.
We index these rules
by $j=1,2,3,4$, in this order. Peer credit uses four
question-disjoint source folds and excludes the response's own fold.
Source advantages use the original response groups.
Here, the source head gradient weights each response by
its inverse valid length.

For valid source positions $t=1,\ldots,L_i$, write
$\mu_{i,t}=\mu^{(-f(i))}_{i,t}$ for the other-fold credit and
$E_{i,t}=\|u_{i,t}\|_2^2$ for the direction's squared norm.
The continuous local gate is
\[
 \xi_{i,t}=\frac{[b_{i,t}]_+}
 {[b_{i,t}]_++\alpha c^0_{i,t}+\epsilon},
\]
with $\alpha=0.0025$ and $\epsilon=10^{-8}$. The candidate sets are
\[
\mathcal I_i^{\rm loc}
 =\{t:b_{i,t}>0,\ E_{i,t}>10^{-24}\},
 \quad \mathcal I_i^{\rm peer}
 =\{t\in\mathcal I_i^{\rm loc}:\mu_{i,t}>0\}.
\]
Let $\mathcal I_i^{\rm top}$ contain the
$|\mathcal I_i^{\rm peer}|$ largest-credit positions in
$\mathcal I_i^{\rm loc}$.
The four allocation scores are
\[
\begin{aligned}
 h^{(1)}_{i,t}&=[b_{i,t}]_+\mathbf1[t\in\mathcal I_i^{\rm loc}],\quad
 h^{(2)}_{i,t}=[b_{i,t}]_+\mathbf1[t\in\mathcal I_i^{\rm top}],\\
 h^{(3)}_{i,t}=h^{(4)}_{i,t}
 &=[\mu_{i,t}]_+\xi_{i,t}\mathbf1[t\in\mathcal I_i^{\rm peer}].
\end{aligned}
\]
\par\Needspace{7\baselineskip}
Their costs and common physical logit-energy budget are
\[
\begin{aligned}
 C^{(j)}_{i,t}&=
 \begin{cases}
 E_{i,t},&j=1,2,3,\\
 E_{i,t}(1+\rho_{{\rm diag},i,t}),&j=4,
 \end{cases}\qquad
 B_i^{\rm phys}=\sum_{t=1}^{L_i}E_{i,t}(w^0_{i,t})^2,
\end{aligned}
\]
where $w^0$ is the base weight in Eq.~\eqref{eq:app_credit_budget}.
Set $\widetilde w^{(j)}_{i,t}=h^{(j)}_{i,t}/C^{(j)}_{i,t}$ when
$h^{(j)}_{i,t}>0$ and $C^{(j)}_{i,t}>10^{-24}$, and zero otherwise.
If the budget or the denominator below is zero, all response weights
for that rule are zero. Otherwise, define
\[
 w^{(j)}_{i,t}
 =\widetilde w^{(j)}_{i,t}
 \sqrt{\frac{B_i^{\rm phys}}
 {\sum_{s=1}^{L_i}E_{i,s}(\widetilde w^{(j)}_{i,s})^2}}.
\]
Each nonzero allocation spends the same physical logit energy per
response. This diagnostic averages responses equally and uses physical
energy rather than the risk-weighted budget of the training objective.

\Needspace{6\baselineskip}
Writing $a_{i,t}$ for the fixed hidden representation, the four-row
virtual head change is
\begin{equation}
 \Delta W^{(j)}_{\mathcal C}
 =\frac1{32}\sum_{i=1}^{32}\frac1{L_i}
   \sum_{t=1}^{L_i}w^{(j)}_{i,t}\,(u_{i,t})_{\mathcal C}a_{i,t}^{\top}.
 \label{eq:app_virtual_head}
\end{equation}
For a calibration or evaluation question $q$, let $a_q$ denote the
hidden representation at its fixed short answer prefix and define
$\Delta z_q^{(j)}=\Delta W^{(j)}_{\mathcal C}a_q$.
Using the centered RMS $s(\cdot)$ in Eq.~\eqref{eq:app_choice_rms}, the
calibration set $\mathcal Q_{\rm cal}$ determines
\[
 \sigma_j^2=\frac1{16}\sum_{q\in\mathcal Q_{\rm cal}}
 s(\Delta z_q^{(j)})^2.
\]
We require $\sigma_j>10^{-12}$; all four rules satisfy this condition.
The evaluation logits are
\[
 z'_{q,\mathcal C}=z_{q,\mathcal C}
 +\frac{\eta}{\sigma_j}\Delta z_q^{(j)}.
\]
Perturbations have centered RMS $\eta$ over calibration logits.

Figure~\ref{fig:app_virtual_transfer} compares the mean virtual transfer
at the primary strength and the question-level entropy and log-odds
changes under peer plus risk. Answer correctness is preserved at this
strength, allowing a focused comparison of distributional changes.
Paired entropy and gold log-odds readouts resolve how local updates
redistribute confidence among answers.

\begin{figure}[tb]
\centering
\includegraphics[width=\textwidth]{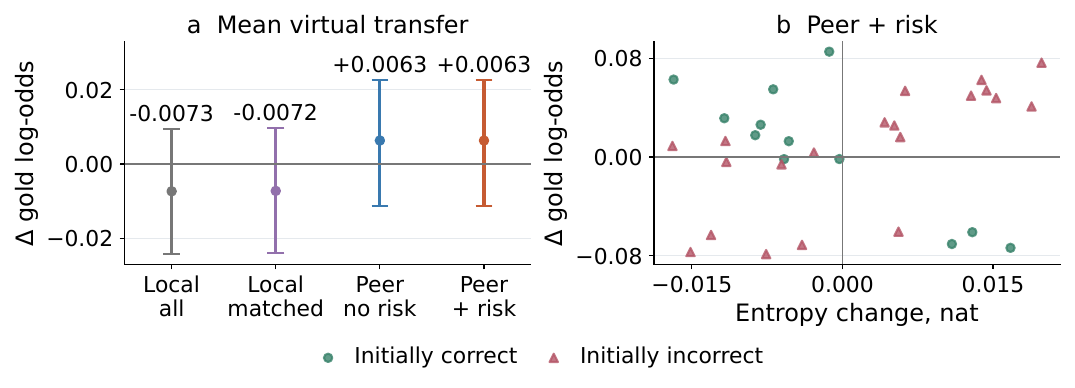}
\caption{\textbf{Virtual output-head transfer at strength 0.03.}
Panel a compares allocation rules by mean gold-option log-odds change;
bars give 95\% task-stratified paired question-bootstrap intervals.
Panel b relates entropy and gold-option log-odds changes under peer
plus risk. Green circles and red triangles denote initially correct
and incorrect predictions, respectively.}
\label{fig:app_virtual_transfer}
\end{figure}

\subsection{Additional Probe Strengths}
Figure~\ref{fig:app_probe_strengths} compares all three prespecified
strengths using the same fixed samples. The primary strength remains
0.03; the additional settings characterize the strength dependence of
conditional answer readouts under this protocol.

\endgroup
\begin{figure}[tb]
\centering
\includegraphics[width=\textwidth]{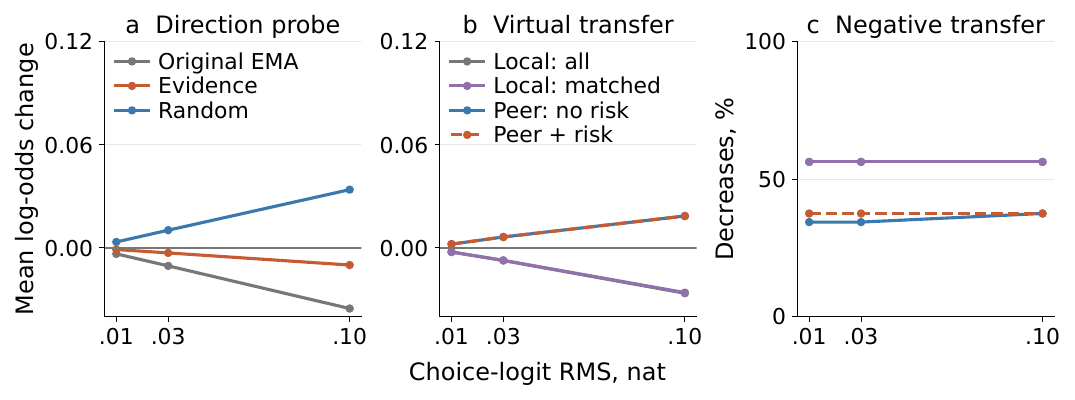}
\caption{\textbf{Sensitivity to perturbation strength.}
Panel a averages direction-probe changes over source questions;
panels b and c evaluate transfer on held-out questions. Colors identify
directions or allocation rules. Dashed orange denotes peer plus risk,
which overlaps peer without risk. Lines join measured strengths.}
\label{fig:app_probe_strengths}
\end{figure}

\Needspace{8\baselineskip}
\section{Computational Cost}
\label{sec:app_cost}

\begin{table}[tb]
\centering\small
\caption{\textbf{Training cost.} Median and upper-percentile step times,
total GPU-hours, and peak memory for the six-method comparison.
Memory columns report the largest worker-averaged allocated and reserved
peaks during training. Measurement scope is described below.}
\label{tab:app_cost}
\setlength{\tabcolsep}{4pt}
\begin{tabular*}{\linewidth}{@{\extracolsep{\fill}}lrrrrr@{}}
\toprule
Method & Median, s & 90th percentile, s & GPU-hours & Allocated, GiB & Reserved, GiB \\
\midrule
GRPO~\citep{ref15} & 62.88 & 76.42 & 14.76 & 16.82 & 26.43 \\
PAPO~\citep{ref56} & 75.34 & 87.38 & 17.53 & 17.95 & 26.71 \\
VGPO~\citep{ref79} & 64.08 & 81.45 & 15.46 & 17.35 & 26.57 \\
TPAE~\citep{ref81} & 84.72 & 99.78 & 20.02 & 17.97 & 30.18 \\
CFPO~\citep{ref80} & 82.79 & 98.51 & 19.35 & 17.42 & 25.77 \\
RAVEC & 119.12 & 142.60 & 27.89 & 22.05 & 46.86 \\
\bottomrule
\end{tabular*}
\end{table}

Table~\ref{tab:app_cost} compares training cost on four NVIDIA H20-3e
GPUs. Timing covers generation, reward computation, teacher and visual
correction calculations, and optimization. It includes validation and
saving within training steps.

GPU-hours equal the device count times the accumulated
step duration; initialization and queue waiting are excluded.
\Needspace{6\baselineskip}
\section{Case Studies and Input Prompts}
\label{app:cases}

\subsection{Token-Aligned Paired Examples}
\begin{figure}[H]
\centering
\includegraphics[width=.96\linewidth,height=0.80\textheight,keepaspectratio]{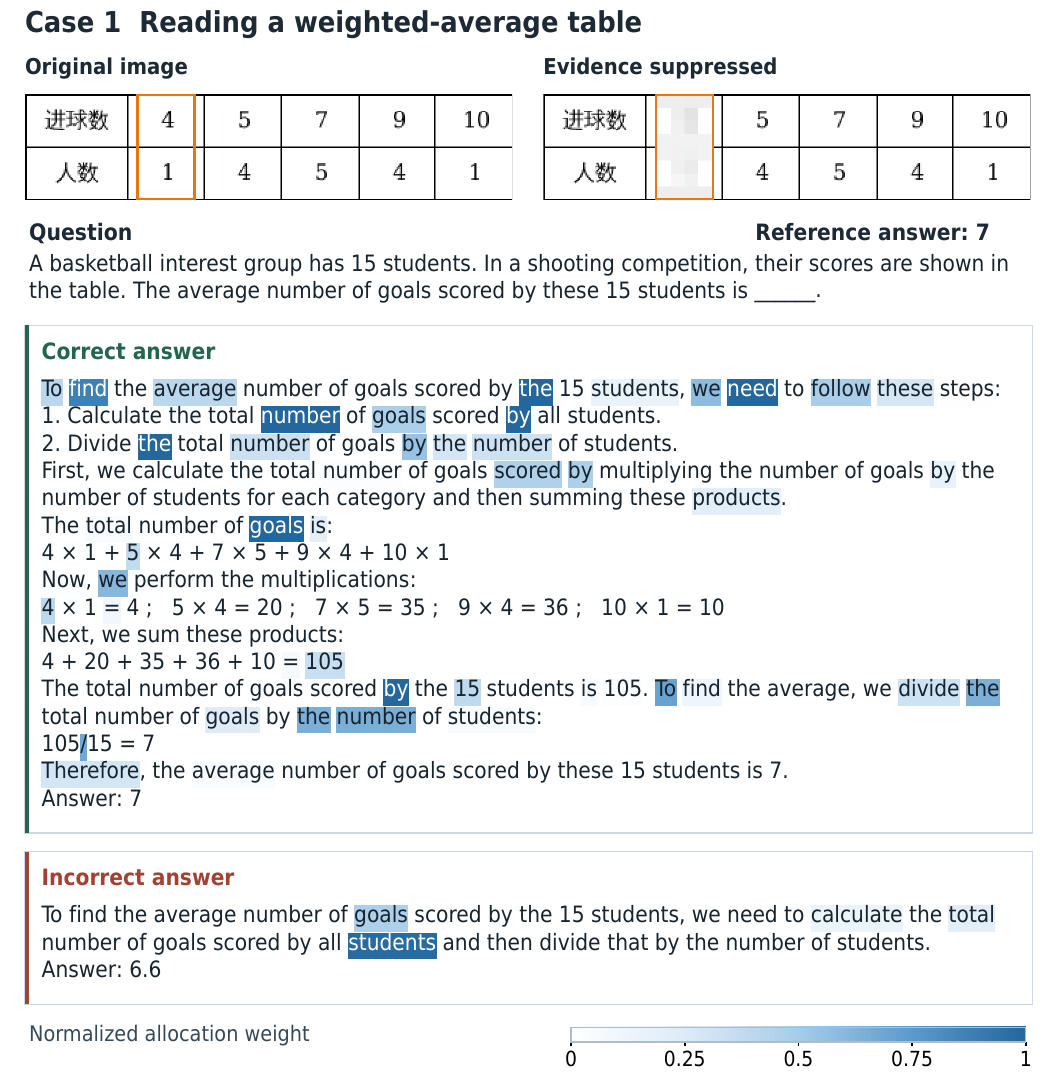}
\caption{\textbf{Reading a weighted-average table.}
The orange rectangle covers the first numerical column: four goals scored
by one student. Pixelation makes both entries unreadable. The correct response computes a weighted average, while the incorrect
response omits the calculation. Blue intensity shows normalized allocation.}
\label{fig:app_case1}
\end{figure}

\begin{figure}[H]
\centering
\includegraphics[width=\linewidth,height=0.80\textheight,keepaspectratio]{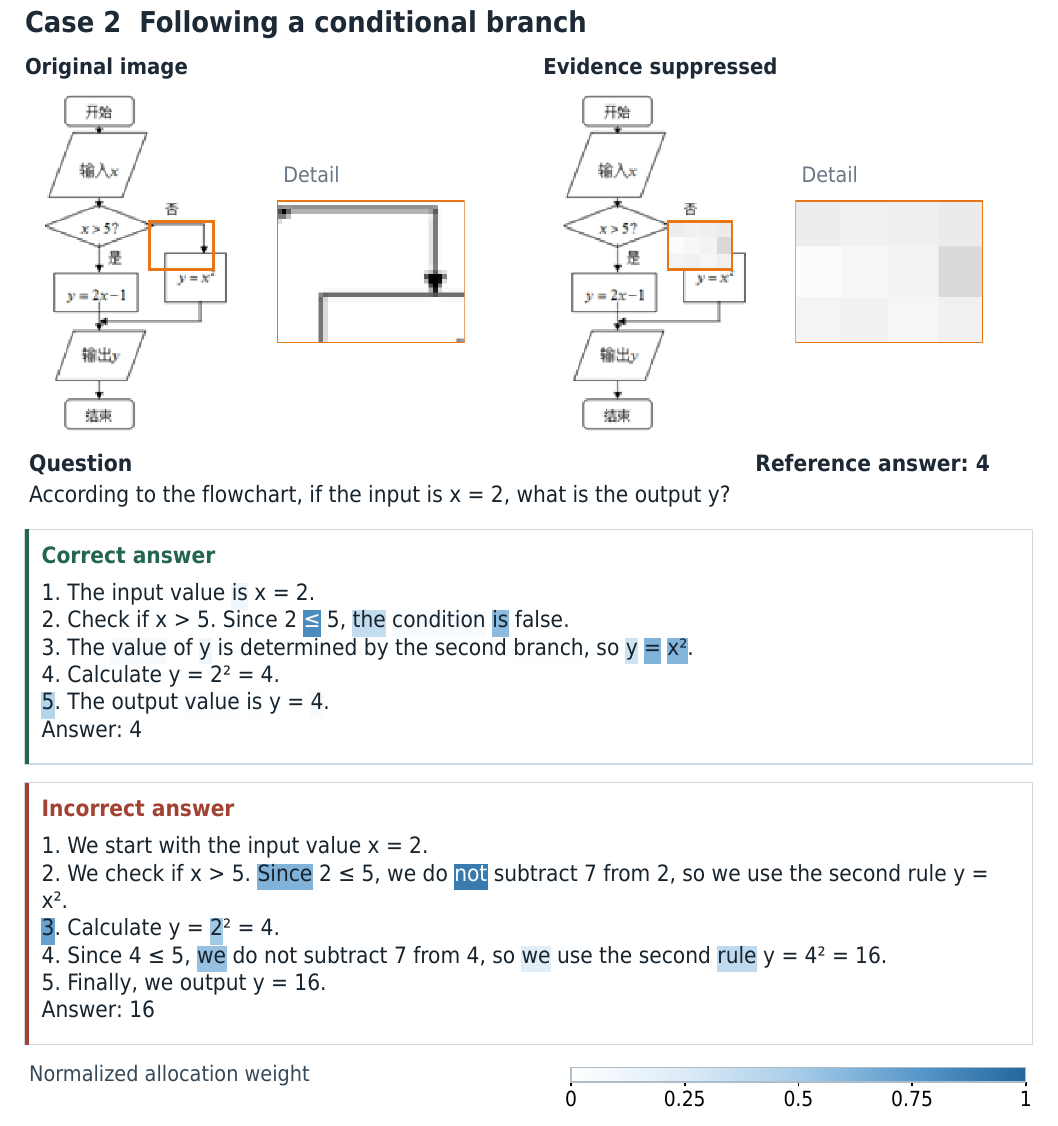}
\caption{\textbf{Following a conditional branch.}
The orange rectangle and enlarged details show the connector into the
``no'' branch before and after pixelation. The branch equation remains
visible. The correct response applies $y=x^2$ once; the incorrect response repeats
it. Allocation colors use the shared scale in Figure~\ref{fig:app_case1}.}
\label{fig:app_case2}
\end{figure}

\begin{figure}[H]
\centering
\includegraphics[width=.88\linewidth,height=0.80\textheight,keepaspectratio]{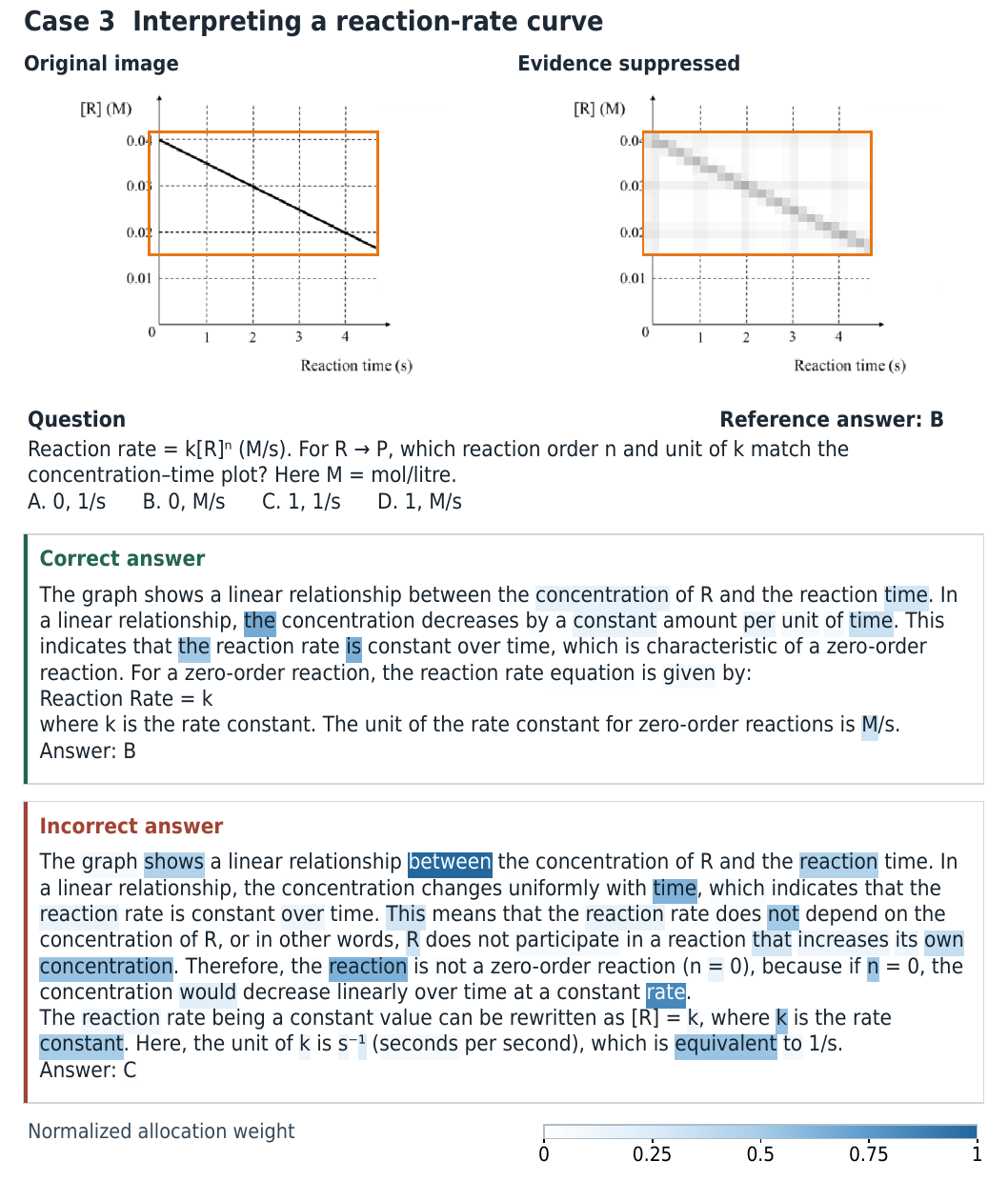}
\caption{\textbf{Visual correction within reaction-rate reasoning.}
The orange rectangle covers the concentration--time curve. Pixelation
retains its coarse linear trend while suppressing local detail. The
correct response relates a zero-order reaction to rate-constant units
M/s. Both responses receive visual correction within their curve
interpretation and unit analysis. The shared color scale makes
token-level allocation comparable across the two reasoning paths.}
\label{fig:app_case3}
\end{figure}

Figures~\ref{fig:app_case1}--\ref{fig:app_case3} compare selected correct
and incorrect responses from Section~\ref{app:paired_allocation}.
The evidence rectangles cover relevant image content, and both responses
receive nonzero content weights. Each figure shows the original image
and its localized pixelation. Responses retain their substantive content
and errors; control markers are hidden. The questions are shortened
where needed for readability.

Let $\bar w_g$ be the mean calibrated weight of the original tokens
contributing to each visible content group. Control markers, layout
syntax, and sentence punctuation are excluded. With $q_{.99}$ denoting
the 99th percentile of group means across paired responses, including
zeros, the shared
color scale is
\[
 c_g=\min\!\left\{\frac{\log(1+\bar w_g)}{\log(1+q_{.99})},\,1\right\}.
\]
White denotes zero and blue intensity increases with allocation.
Saturation affects display only; all responses share the same scale.
Colors consistently encode allocated correction strength across examples.

The first two cases illustrate numerical aggregation in a table and
selection of a branch in a diagram. The third traces visual correction
through the interpretation of a reaction-rate curve and its units.
The shared color scale connects correction strength to the displayed
content, enabling inspection of allocation alongside the
reasoning in each response.

\subsection{Literal Input Prompts}
\label{sec:app_prompts}
Figures~\ref{fig:app_prompt_p1a}--\ref{fig:app_prompt_example} reproduce
the instructions and message roles, with example-specific
placeholders. Policy prompts P2 and P3 provide images and a question without
the reference answer or annotation reasoning. Annotation prompt P1
provides the standard answer to identify a localized evidence region.
Unlocalizable examples supply no region.

Probe prompt P4 retains the source reasoning up to the final answer
letter; calibration and evaluation instead use a shared short prefix.
Source reasoning can reveal the answer, as Figure~\ref{fig:app_prompt_example} shows;
option probabilities therefore reflect both visual and textual information.

\begin{figure}[H]
\centering
\setlength{\fboxsep}{6pt}
\fbox{\begin{minipage}{\dimexpr\textwidth-2\fboxsep-2\fboxrule\relax}
\small\textbf{P1. Evidence annotation: system message}\par\smallskip
\textbf{System}\par
{\small\ttfamily\raggedright
You annotate visual evidence for multimodal reasoning training. Inspect every image in order,\allowbreak{}\par
the unchanged problem,\allowbreak{} and its gold answer. Locate one tight region that directly contributes to the answer\par
and write a concise image-grounded reasoning process. The rectangle must isolate a specific object,\allowbreak{} label,\allowbreak{}\par
number,\allowbreak{} or small visual relation; never choose the entire image or a broad half-image. For counting,\allowbreak{}\par
isolate one counted item whose removal would change the count,\allowbreak{} not the whole collection. For comparisons,\allowbreak{}\par
isolate one of the compared values or objects. The rectangle area must be at most 35\% of that image.\par
If no answer-contributing local region can be identified reliably,\allowbreak{} mark the sample unlocalizable.\par
Your response MUST begin with \{ and contain a single JSON object. Do not write analysis before it.\par
Return JSON only,\allowbreak{} with exactly:\allowbreak{}\par
\{"status":\allowbreak{}"annotated",\allowbreak{}"evidence\_image\_index":\allowbreak{}0,\allowbreak{}"drop\_bbox":\allowbreak{}[left,\allowbreak{}top,\allowbreak{}right,\allowbreak{}bottom],\allowbreak{}"reasoning":\allowbreak{}"..."\}\par
Coordinates are integers normalized to 0..1000 and the image index is zero-based. Reasoning must not contain\par
\textless{}think\textgreater{},\allowbreak{} \textless{}/think\textgreater{},\allowbreak{} \textbackslash{}boxed,\allowbreak{} or a separately generated final answer. If no reliable local evidence exists return\par
\{"status":\allowbreak{}"unlocalizable",\allowbreak{}"reason":\allowbreak{}"..."\}.\par}
\smallskip
\end{minipage}}
\caption{\textbf{Evidence annotation prompt.} The annotator receives the images, question, and standard answer and either identifies a localized region or marks the example unlocalizable.}
\label{fig:app_prompt_p1a}
\end{figure}
\begin{figure}[H]
\centering
\setlength{\fboxsep}{6pt}
\fbox{\begin{minipage}{\dimexpr\textwidth-2\fboxsep-2\fboxrule\relax}
\small\textbf{P1. Annotation input and retry}\par\smallskip
\textbf{User}\par
{\small\ttfamily\raggedright
[images in their original order]\par
Problem:\allowbreak{}\par
[unchanged problem]\par
\mbox{}\par
Gold answer:\allowbreak{}\par
[standard answer]\par}
\smallskip
\textbf{Retry after invalid output}\par
{\small\ttfamily\raggedright
Your previous annotation was invalid:\allowbreak{} [first 180 characters of validation feedback]\par
Choose a tight box around ONE answer-contributing object,\allowbreak{} numeral,\allowbreak{} or label. For a counting question,\allowbreak{} choose ONE counted item. If impossible,\allowbreak{} return the unlocalizable JSON object. Start directly with \{.\par}
\smallskip
\end{minipage}}
\caption{\textbf{Annotation input and retry prompt.} Placeholders denote example-specific material. Feedback is added after invalid output. The standard answer is used only for annotation.}
\label{fig:app_prompt_p1b}
\end{figure}
\begin{figure}[H]
\centering
\begin{minipage}[t]{.49\linewidth}\vspace{0pt}
\centering
\setlength{\fboxsep}{6pt}
\fbox{\begin{minipage}{\dimexpr\textwidth-2\fboxsep-2\fboxrule\relax}
\small\textbf{P2--P3. Policy training and evaluation}\par\smallskip
\textbf{System}\par
{\small\ttfamily\raggedright
You are a helpful assistant.\par}
\smallskip
\textbf{User}\par
{\small\ttfamily\raggedright
[original question,\allowbreak{} with images at their original positions and any answer options]\par
\mbox{}\par
You first think through the reasoning process as an internal monologue,\allowbreak{} enclosed within \textless{}think\textgreater{} \textless{}/think\textgreater{} tags. Then,\allowbreak{} provide your final answer enclosed within \textbackslash{}boxed\{\}.\par}
\smallskip
\end{minipage}}
\caption{\textbf{Shared policy prompt.} Training and evaluation use questions and images without standard answers or annotation reasoning.}
\label{fig:app_prompt_p2p3}
\end{minipage}\hfill
\begin{minipage}[t]{.49\linewidth}\vspace{0pt}
\nextfloat
\centering
\setlength{\fboxsep}{6pt}
\fbox{\begin{minipage}{\dimexpr\textwidth-2\fboxsep-2\fboxrule\relax}
\small\textbf{P4. Fixed-prefix mechanism probes}\par\smallskip
\textbf{System}\par
{\small\ttfamily\raggedright
You are a helpful assistant.\par}
\smallskip
\textbf{User}\par
{\small\ttfamily\raggedright
[original question,\allowbreak{} with images at their original positions and any answer options]\par
\mbox{}\par
You first think through the reasoning process as an internal monologue,\allowbreak{} enclosed within \textless{}think\textgreater{} \textless{}/think\textgreater{} tags. Then,\allowbreak{} provide your final answer enclosed within \textbackslash{}boxed\{\}.\par}
\smallskip
\textbf{Assistant continuation for calibration and evaluation}\par
{\small\ttfamily\raggedright
\textless{}think\textgreater{}\par
\textless{}/think\textgreater{}\par
\textbackslash{}boxed\{\par}
\smallskip
\end{minipage}}
\caption{\textbf{Answer-prefix probe.} Calibration and evaluation questions share the displayed prefix. Source questions use their generated reasoning up to the answer letter, as in Figure~\ref{fig:app_prompt_example}. Option probabilities are read over A--D.}
\label{fig:app_prompt_p4}
\end{minipage}
\end{figure}

\begin{figure}[H]
\centering
\setlength{\fboxsep}{6pt}
\fbox{\begin{minipage}{\dimexpr\textwidth-2\fboxsep-2\fboxrule\relax}
\small\textbf{P4. Example response prefix}\par\smallskip
\textbf{User question; policy instruction as in P2}\par
{\small\ttfamily\raggedright
\textless{}image\textgreater{}The three angles of a triangle are shown in the figure. What type of triangle is this? \par
Please select the correct option:\allowbreak{} A. Right-angled; B. Acute-angled; C. Obtuse-angled; D. No correct answer\par}
\smallskip
\textbf{Assistant prefix, ending before answer A}\par
{\small\ttfamily\raggedright
\textless{}think\textgreater{}\par
The sum of the angles in any triangle is always 180 degrees. Let's check the angles provided:\allowbreak{}\par
- One angle is 90 degrees (a right angle).\par
- Another angle is 50 degrees.\par
- The third angle is 40 degrees.\par
\mbox{}\par
Now,\allowbreak{} let's add them up:\allowbreak{} 90 + 50 + 40 = 180 degrees,\allowbreak{} which confirms that these are the angles of a triangle. \par
\mbox{}\par
Since one of the angles is a right angle (90 degrees),\allowbreak{} this triangle is a right-angled triangle.\par
\textless{}/think\textgreater{}\par
\textbackslash{}boxed\{\par}
\smallskip
\end{minipage}}
\caption{\textbf{Example response prefix.} The reasoning already identifies a right angle, so answer probabilities condition on both visual evidence and preceding text.}
\label{fig:app_prompt_example}
\end{figure}

\end{document}